\documentclass[pmlr,twocolumn]{jmlr}
\usepackage{multirow}
\usepackage{longtable}
\usepackage{booktabs}
\usepackage{siunitx}

\usepackage{todonotes}

\newcommand{\xhdr}[1]{\vspace{1.7mm}\noindent{{\bf #1.}}}

\usepackage{enumitem}
\usepackage{amsmath}
\usepackage{bbm}

\DeclareMathOperator*{\argmin}{arg\,min}

\usepackage[compact]{titlesec}

\ifdefined \remark 
\else
\newtheorem{remark}{Remark}
\fi

\theorembodyfont{\upshape}
\theoremheaderfont{\scshape}
\theorempostheader{:}
\theoremsep{\newline}

\jmlrvolume{193}
\firstpageno{1}

\jmlryear{2022}
\jmlrworkshop{Machine Learning for Health (ML4H) 2022}

\editor{}

\author{}

\begin{document}
\title[TB Adherence]{Predicting Treatment Adherence of Tuberculosis Patients at Scale}

 
 \author{
  \Name{Mihir Kulkarni\nametag{\thanks{Equally contributing first author.}}}$^1$ \Email{mihir@wadhwaniai.org}\\
  \Name{Satvik Golechha\nametag{\footnotemark[1]}}$^1$
  \Email{satvik@wadhwaniai.org}\\
  \Name{Rishi Raj\nametag{\footnotemark[1]}}$^1$
  \Email{rishi@wadhwaniai.org}\\
  \Name{Jithin K.\ Sreedharan\nametag{\footnotemark[1]}}$^2$
  \Email{jithin@cse.iitk.ac.in}\\
  \Name{Ankit Bhardwaj\nametag{\thanks{Work done at Wadhwani AI.}}}$^3$ \Email{bhardwaj.ankit@nyu.edu}\\
  \Name{Santanu Rathod}$^1$
  \Email{santanu@wadhwaniai.org}\\
  \Name{Bhavin Vadera}$^4$
  \Email{bvadera@usaid.gov}\\
  \Name{Jayakrishna Kurada}$^1$
  \Email{jayakrishna@wadhwaniai.org}\\
  \Name{Sanjay Mattoo}$^5$
  \Email{mattoos@rntcp.org}\\
  \Name{Rajendra Joshi}$^5$
  \Email{ddgtb@rntcp.org}\\
  \Name{Kirankumar Rade}$^6$
  \Email{radek@who.int}\\
  \Name{Alpan Raval}$^1$ 
  \Email{alpan@wadhwaniai.org}\\
\addr $^1$Wadhwani Institute for Artificial Intelligence (Wadhwani AI) $^2$IIT Kanpur $^3$Courant Institute of Mathematical Sciences, New York University $^4$USAID $^5$Central TB Division (India) $^6$World Health Organization
 }
\maketitle


\begin{abstract}
Tuberculosis (TB), an infectious bacterial disease, is a significant cause of death, especially in low-income countries, with an estimated ten million new cases reported globally in $2020$. 
While TB is treatable, non-adherence to the medication regimen is a significant cause of morbidity and mortality. 
Thus, proactively identifying patients at risk of dropping off their medication regimen enables corrective measures to mitigate adverse outcomes. 
Using a proxy measure of extreme non-adherence and a dataset of nearly $700,000$ patients from four states in India, we formulate and solve the machine learning (ML) problem of early prediction of non-adherence based on a custom rank-based metric. 
We train ML models and evaluate against baselines, achieving a $\sim 100\%$ lift over rule-based baselines and $\sim 214\%$ over a random classifier, taking into account country-wide large-scale future deployment. 
We deal with various issues in the process, including data quality, high-cardinality categorical data, low target prevalence, distribution shift, variation across cohorts, algorithmic fairness, and the need for robustness and explainability. Our findings indicate that risk stratification of non-adherent patients is a viable, deployable-at-scale ML solution.

As the official AI partner of India's Central TB Division, we are working on multiple city and state-level pilots with the goal of pan-India deployment.
\end{abstract}

\section{Introduction and Problem Statement}
\label{sec:intro}
Tuberculosis is one of the world's great scourges; 
it was the cause of the second-highest number of deaths from a single infectious agent in $2020$, with an estimated $10$ million cases worldwide, and $1.5$ million deaths~\citep{WHO_TB_2021}.
In India, with a median estimated incidence of $2.59$ million TB cases in $2020$, TB is a serious public health concern~\citep{WHO_TB_2021}.
While drug-sensitive TB (DSTB) is treatable through a standard drug combination typically administered for six months, drug-resistant TB (DRTB) is much more difficult to treat. A major cause of DRTB is lack of adherence to the standard treatment~\citep{volmink2007directly, jain2008multidrug, mekonnen2018non}. In order to drive patient adherence to the treatment regimen, 
Directly Observed Treatment Short-course (DOTS), or direct observation of the patient, has so far been the mainstay recommendation for successful TB treatment outcomes~\citep{INDIA_TB_REPORT}. 

Non-adherence to TB treatment exists in many forms; an extreme form is loss to follow-up (LFU), representing a treatment interruption 
of thirty consecutive days or more. In $2020$, India's national TB elimination program (NTEP) reported that $\sim 3$\% of TB patients who started treatment became LFU, with up to 13\% LFU rates in DRTB patients ~\citep{INDIA_TB_REPORT}. 
LFU patients pose serious challenges as silent transmitters of TB and contribute to increased risk of morbidity, mortality, and cost burden~\citep{SALLA, osman2021early, zheng2020insufficient}.

While rigorous DOTS monitoring and periodic patient home visits are core mandates for TB field staff in India ~\citep{INDIA_TB_REPORT}, providing the same level of intensive monitoring to all patients is infeasible in the resource-constrained TB healthcare setting, leading to LFU. Other factors causing high LFU include alcoholism, smoking addictions, previous history of TB, adverse drug reactions, chronicity of treatment and pill burden, lack of family/social support, migrant status, out-of-pocket expenditure, stigma, and poor knowledge of the disease~\citep{jaggarajamma2007reasons, washington2020differentiated}. 

\xhdr{Problem Statement}
 We model the task of predicting whether a patient $p$ will be LFU as a binary classification problem with target label $y_p$ ($1$ for LFU and $0$ for non-LFU), where patient-level features are represented as $\mathbf{x}_p = \{x_{p1}, x_{p2}, \ldots , x_{pd}\}$. The model outputs a risk score $\hat{y}_p = f(\mathbf{x_p})$, corresponding to the propensity for a patient becoming LFU, where $f: \mathbb{R}^{d} \xrightarrow{} [0, 1]$.
 Features of $n$ patients can be collected into a $n \times d$ feature matrix $\mathbf{X}$. True and predicted label vectors are denoted as $\mathbf{y}$ and $\mathbf{\hat{y}}$, respectively.

\begin{figure*}[t]
\floatconts
  {fig:journey}
  {  \vspace{-2 em}
  \caption{Proposed patient journey, from being diagnosed TB positive to receiving differentiated care based on the risk score computed by our solution.
  \vspace{-1 em}
  }
  }
  {\includegraphics[width=\linewidth]{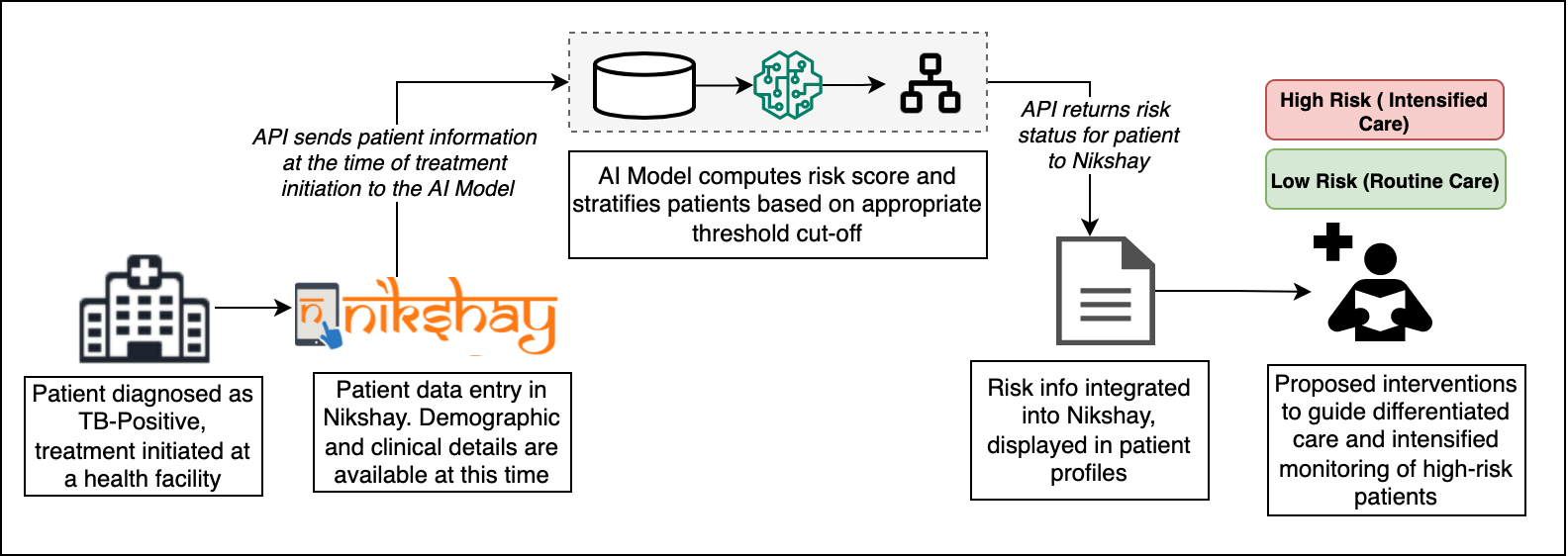}}
  \vspace{-1.5 em}
\end{figure*}

Our solution is a machine learning (ML) tool that recommends early stratification of TB patients at risk of LFU to enable effective differentiated interventions and care, through modified patient monitoring workflows and operating procedures.
Risk stratification leads to optimal usage of existing resources and enables systematic prioritization of patients for intensified care. \autoref{fig:journey} shows how the solution fits into the journey of a patient registered in Nikshay, a web-enabled patient management system for TB control run by India's National Tuberculosis Elimination Program (NTEP). 

\xhdr{Contributions}
The major contributions of our work are as follows:

    \noindent 1.\ {\em Problem formulation}: We formulate and solve an ML problem to combat a large-scale public health problem affecting millions of patients worldwide, through guiding benefit allocation in a resource-constrained setting. 
    Deployment considerations and real-world constraints are incorporated into the model to maximize the solution's effectiveness. These include a custom rank-based sensitivity metric for evaluation and model selection, and splitting the data temporally (forward splitting) to mimic the deployment scenario directly.
    
    \noindent 2.\ {\em Model enhancements}: The dataset has a large size, low target prevalence, a substantial number of high-cardinality categorical features, and is subject to data distribution shifts due to seasonal variations and a diverted focus of the healthcare system on other epidemics. We extensively search for the best encoding-model pair in this scenario, including state-of-the-art categorical variable encodings, tabular deep learning models, and fully interpretable boosting algorithms. Other techniques that we experiment with include data augmentation, ensembling, interpretability, and algorithmic fairness, enabling us to guide deployment and convince public health authorities.
    \noindent 3.\ {\textit{Evaluation and generalization capability}}: We set aside six months of data for a passive evaluation to incorporate most out-of-distribution cases. We comprehensively evaluate models across multiple cohorts of interest -- geographical, gender-based, type of TB, and public/private health facility. We see robust generalization across time and geographies and provide good measures of predictive multiplicity~\citep{PRED_MULT} as evidence of the robustness of our model scores.

\noindent 4. {\textit{Potential social impact}}:
The goal of the project is to systematically deploy the solution across all $780$ districts in India, impacting over a million TB patients annually. 
At each geographical cohort, health workers will monitor a dashboard that displays the risk status for each patient, which field staff will use to provide prioritized interventions. 
This will allow targeted utilization of health workers' bandwidth. For $325,190$ patients with treatment initiated in the last six months of $2020$, our ML solution leads to a potential impact on $5587$ lives, compared to $1808$ with random targeting and $2974$ with the best rule-based baseline model (\autoref{sec:evaluation}). 
We are now working closely with the Central TB Division, Govt.\ of India, and several Indian states for a pilot deployment. 

\section{Related Work}

There has been significant work on predicting adherence to medication regimens, for HIV/AIDS \citep{ettenhofer2010reciprocal}, heart failure \citep{davis2014pharmacist}, and mental disorders \citep{adams2000predicting}. Explainable Boosting Machines (EBMs) ~\citep{GAM} have been used in interpretable healthcare settings for predicting adverse outcomes in pregnancy \citep{bosschieter2022using} and mental health \citep{lee2021micromodels}. 

There has also been work done on analyzing risk factors and building statistical or machine-learning tools to assist TB care. Risk factors for TB have been studied in Tajikistan \citep{rahman2021deep} and Papua New Guinea \citep{umo2021factors}. Statistical models such as survival analysis or regression have been used across the developing world~\citep{shargie2007determinants, kliiman2010predictors, roy2015risk, xu2017detection}. XGBoost models \citep{XGBOOST} with deep learning-based feature extraction have been used on chest X-rays \citep{wohlleben2017risk}. In the Indian context, \citep{killian2019learning} construct ML models to predict TB medication adherence patterns using past data recorded through the 99DOTS \citep{cross201999dots} tracking system, an approximate  adherence measure, using data of $\sim 17,000$ patients from Mumbai.  

While we also leverage gradient-boosted tree models such as EBMs and XGBoost for adherence prediction, our work directly uses the NTEP program data at scale, culled from across locations in a country that has the highest number of TB-afflicted individuals in the world. The data is then used to predict true non-adherence, albeit of an extreme kind, rather than an approximate proxy.

\section{Data}
\label{sec:data}

\xhdr{Nikshay data}
In this work, we use TB patient data obtained via the Nikshay\footnote{\url{https://www.nikshay.in/Home/AboutUs}} system.
Nikshay is a national patient management system for TB control developed and maintained by the Central TB Division (CTD), Government of India, the National Informatics Centre (NIC), and the World Health Organization (WHO). 

It contains a line list database of TB patients, longitudinally tracking each patient's health status and interactions with the public health system, from diagnosis to treatment completion. 

We employ Nikshay data from $2020$ of four Indian states: Karnataka, Maharashtra, Uttar Pradesh, and West Bengal. The data (\tableref{tab:states}), being highly sensitive and personal, is anonymized and stripped off of personally identifiable information (patient's name, number, address, etc.). The data consists of separate tables called \emph{registers}, linked by a common patient ID. 
The $2020$ data has seven such registers: Notification, Comorbidity, Contact Tracing, Patient Data, Adherence, Patient Lab, and DMC (Designated Microscopy Centre) (\tableref{tab:registers}).
%
Details of the data analysis and pre-processing pipeline are given in Appendix \ref{app:subsec:nikreg}.

\begin{table*}[ht]
\centering
\floatconts
  {tab:states}%
  {\caption{\small State-wise data distribution, split into data for modeling and data for passive evaluation, with number of patients and LFU prevalence.}
  \vspace{-2 em}
  }%
  {%
\resizebox{\textwidth}{!} 
{
\begin{tabular}{ccccccc}
\toprule
 \multicolumn{3}{c}{\textbf{{} {} {} {} {} {} {} {} {} {} {} {} {} {} {} {} {} {} {} {} {} {} {} {} {} Overall}}    & \multicolumn{2}{c}{\textbf{Modeling Split}} & \multicolumn{2}{c}{\textbf{Passive Evaluation Split}} \\
\multirow{-2}{*}{\textbf{States}} & \text{No. of Patients} & \text{LFU Rate} & \text{No. of Patients}             & \text{LFU Rate}            & \text{No. of Patients}               & \text{LFU Rate}              \\
\midrule
Karnataka
& 65,120                   & 2.67\%                      & 34,171                               & 2.70\%                       & 28,041                                 & 2.91\%                         \\
Uttar Pradesh  
& 376,028                 & 3.39\%                     & 173,405                             & 4.04\%                       & 183,841                               & 3.11\%                         \\
West Bengal  
& 79,807                   & 2.05\%                 & 39,651                               & 2.17\%                       & 37,378                                 & 2.07\%                         \\
Maharashtra
& 157,997                 & 2.51\%                     & 75,103                               & 2.98\%                       & 75,930                                 & 2.28\%                         \\
\midrule
\text{Total}                                   & \text{678,952}        & \text{2.96\%}    & \text{322,330}                    & \text{3.42\%}              & \text{325,190}                      & \text{2.78\%} \\
\bottomrule
\end{tabular}%
}
}
\end{table*}


\subsection{Forward splitting}
\label{subsec:forward_splitting}
The deployment scenario involves training the model on past data and carrying out inference on future data. In order to simulate this, we forward-split the data in time. We reserve the last six months of data as a \emph{passive evaluation split}, denoted as $(\mathbf{y}^{(\text{pes})}, \mathbf{X}^{(\text{pes})})$, which is only used for final model evaluation and reporting, not for training, optimization, model selection, or internal evaluation. All previous data is referred to as the \emph{modeling split}, $(\mathbf{y}^{(\text{ms})}, \mathbf{X}^{(\text{ms})})$. The modeling split is further divided chronologically in $60\!:\!20\!:\!20$ proportion into train, validation, and test sets  (denoted as $(\mathbf{y}^{(\text{ms,str})}, \mathbf{X}^{(\text{ms, str})})$ with $\text{str} \in \{\text{train, val, test} \}$). The validation set is used, as usual, for hyperparameter optimization within a model class, and the test set is used for selecting the best model class. This best model class is then trained on the entire modeling split prior to its final evaluation on the passive evaluation split.

The $60\!:\!20\!:\!20$ proportion was decided based on an empirical investigation of different splits to achieve a balance between having sufficient data for training (\autoref{sec:training}) 
and cohort-wise evaluation~(\autoref{sec:evaluation}). 

\tableref{tab:states} also summarizes numbers of patients by split. Note a significant drop in the fraction of LFU patients between the modeling and the passive evaluation splits.
This can be attributed to the high burden of COVID-19 on the healthcare system during this time period. We expect our models to face such distribution shifts in deployment as well. See Appendix \ref{app:subsec:generalization} for effects of distribution shifts on model performance, where we show the model is by and large robust to them.

\section{Modeling}
\label{sec:training}

\subsection{Encodings}
\label{sec:encodings}

Except for patient age, all features in $\mathbf{X}$ are categorical, and some with high cardinality. Since most models do not work out-of-the-box on such data, we need to encode the features into a real-valued vector space.

We experiment with a number of encoding techniques: those that do not use label information, such as Normalized Count Encoding, Similarity Encoding \citep{SIMILARITY},
and Entity Embedding \citep{ENTITY},
as well as those that use the label as a prior, such as Target Encoding \citep{TARGET}, Leave-One-Out (LOO) Encoding, and CatBoost Encoding \citep{CATBOOST}. We also experiment with recent techniques such as Gap Encoding \citep{DIRTYCAT} and MinHash Encoding \citep{DIRTYCAT} (to handle high-cardinality).

We compare these encoding techniques (see \autoref{fig:cd_enoder_sel} and \tableref{tab:encodings}) using high-depth XGBoost trees, which are appropriate to use since they train fast and are known to do well out of the box on tabular data.

\subsection{Metrics}
Keeping deployment in mind, the following considerations were found to be relevant to NTEP staff. 
First, the ML solution should capture as many patients as possible who will eventually be LFU (high recall). 
Second, the staff workers who will use our solution can typically target only a certain $k\%$ of the patients for intensive monitoring and interventions due to resource limitations and their involvement in other disease programs. 
Further, there is low prevalence ($\sim 3-4\%$) of the positive class among TB patients.
Thus, most standard classification evaluation metrics are ill-suited to the problem. Instead, we focus on the top $k\%$ patients, aiming to achieve high recall in that set, and work with the following two custom metrics:

\noindent $\bullet~\text{Recall}@k$: Recall for top $k\%$ patients when patients are sorted in descending order of the score given by the model. Such metrics are commonly used in ranking problems~\citep{manning2008introduction}.
Feedback from public health experts suggests that TB health workers can typically target roughly $20\%$ of the patients served by their TB unit for interventions. This led us to report recall at $k=20$ as one of the key metrics.

\noindent $\bullet~\text{AvRecall}(10, 40)$: The average Recall$@k$ for $k \in [10, 40]$, i.e., while targeting the top high risk $10\%$ to $40\%$ of patients, a reasonable range for on-ground health worker resource bandwidth, and a more robust measure of performance on a range of $k$ values.

See Appendix \ref{sec:metric_discussion} for further discussion on our metrics.
\subsection{Models}

Our modeling experiments included simple ML models like decision trees, $k$-nearest neighbors ($k$-NN), and naive Bayes classifiers, as well as more recent variants of gradient-boosted decision tree (GBDT) models and deep learning models. We find that all models except  $k$-NN outperform the rule-based baselines (\autoref{sec:rbb}).

GBDT variants include XGBoost~\citep{XGBOOST}, LightGBM~\citep{LIGHTGBM}, and CatBoost~\citep{CATBOOST}. We also investigated two  deep learning architectures that have shown comparable performance to GBDTs on several tabular datasets, TabNet \citep{TABNET} and the heavily regularized multilayer perceptron (MLP)~\citep{COCKTAILS}.

The experiments in Section~\ref{sec:results} show that a number of models perform nearly equally well (\autoref{tab:models}). In such situations, \citep{RASHOMONSET} suggest that there likely exists a  simpler, fully interpretable model with similar performance; we therefore also implement an Explainable Boosting Machine (EBM), an interpretable, tree-based gradient-boosted generalized additive model.

\xhdr{Hyperparameter tuning and model selection}
We perform hyperparameter tuning, and model and encoder selection using the evaluation metric $\text{AvRecall}(10,40)$ as the loss function $\mathcal{L}$ instead of the training loss.

Our entire hyperparameter tuning and encoder/model selection process can be described as follows:

\begin{align*}
        f^{*} &= \argmin_{f \in \mathcal{F}} \min_{\lambda \in \Lambda_f} \mathcal{L}(\mathbf{y}, f(\mathbf{X}; \lambda, E^{*})), ~\text{s.t.}\\
E^{*} & = \argmin_{E \in \mathcal{E}} \min_{\lambda \in \Lambda_{f_{\text{xgb}}}} \mathcal{L}(\mathbf{y}, f_{\text{xgb}}(\mathbf{X}; \lambda, E)), \nonumber
\end{align*}

where $\mathcal{F}$, $\Lambda_f$, and $\mathcal{E}$ denote the set of all models, hyperparameter configurations of model $f$, and encodings we considered, respectively. 
We find that the performance of models depends strongly on the encoding of categorical variables, as in \citep{cerda2020encoding}. With a slight abuse of notation, therefore, we change $f(\mathbf{x})$ to $f(\mathbf{x}; \lambda, E)$, where $\lambda$ and $E$ denote model hyperparameters and encoding respectively. While a joint optimization over $\mathcal{F}$ and $\mathcal{E}$ would be ideal, this is practically infeasible due to dataset size and the combinatorial explosion of possibilities. Hence, we initially perform a best encoder search with the XGBoost ($f_{\text{xgb}}$) model, a well-performing GBDT model on tabular data.
There is also a manual search over hyperparameters of encoders which is not indicated in the expression.

Optimal model hyperparameters are found via a search over $\lambda \in \Lambda_f$, using a Tree-structured Parzen Estimator approach (TPE) \citep{TPE}, as implemented in the Hyperopt library \citep{HYPEROPT}. Appendix~\ref{app:subsec:hyperparam_details} lists model details, including hyperparameters and their ranges.
We used the validation set $\mathbf{X}^{(\text{ms}, \text{val})}$ of the modeling split for hyperparameter optimization, and the corresponding test set $\mathbf{X}^{(\text{ms}, \text{test})}$ for the encoder and model search.
Once the best model $f^*$ and its best hyperparameter set $\lambda_f^*$ are found, the model is retrained on the entire $\mathbf{X}^{(\text{ms})}$ set to yield the model  $f^{(\text{ms})^*}$. [Note that the best model search also includes an ensemble of five other best models; see Appendix~\ref{app:subsec:ensembling} for details.] For final prediction, we apply $f^{(\text{ms})^*}$  to the passive evaluation set as
\[\mathbf{y}_i^{(\text{pes})} = f^{(\text{ms})^*}(\mathbf{x}^{(\text{pes})}_i; \lambda^*_f, E^*),\]

\noindent or on various cohorts as detailed in Section~\ref{sec:evaluation}.

\section{Results and Evaluation}
\label{sec:results}

\subsection{Rule-based baselines}
\label{sec:rbb}
We consider three rule-based heuristics as benchmarks to compare the performance of our ML models. These rules are based on correlations in the data, findings by the CTD, protocols suggested by the Karnataka Health Promotion Trust (KHPT) \citep{khpt}, the TB ground staff, and previous literature vetted by domain experts \citep{cherkaoui2014treatment, jaggarajamma2007reasons, bhatnagar2019user, bhattacharya2018barriers}. These baselines are occasionally used by health workers to prioritize interventions.
We list the baselines with their associated features and performance in Appendix~\ref{app:subsec:rule-based-baselines}. Overall, we find that even the best rule-based baseline suffers from poor LFU recall.

\subsection{Statistical comparisons across models and encodings}
\label{sec:statcomp}

\begin{figure*}[htb!]
\floatconts
  {fig:cd_all}
  {  \vspace{-1.5 em}
\caption{\small 
{\bf Critical difference diagrams} with a Wilcoxon significance analysis on $1000$ bootstrap samples. The numbers on the top line indicate average ranks, and connected ranks via a bold bar indicate that performances are not signiﬁcantly different ($p > 0.05$).}
\vspace{-2 em}
}
  {%
    \subfigure[\small Selecting the best encoder.]{\label{fig:cd_enoder_sel}%
      \includegraphics[scale=0.42]{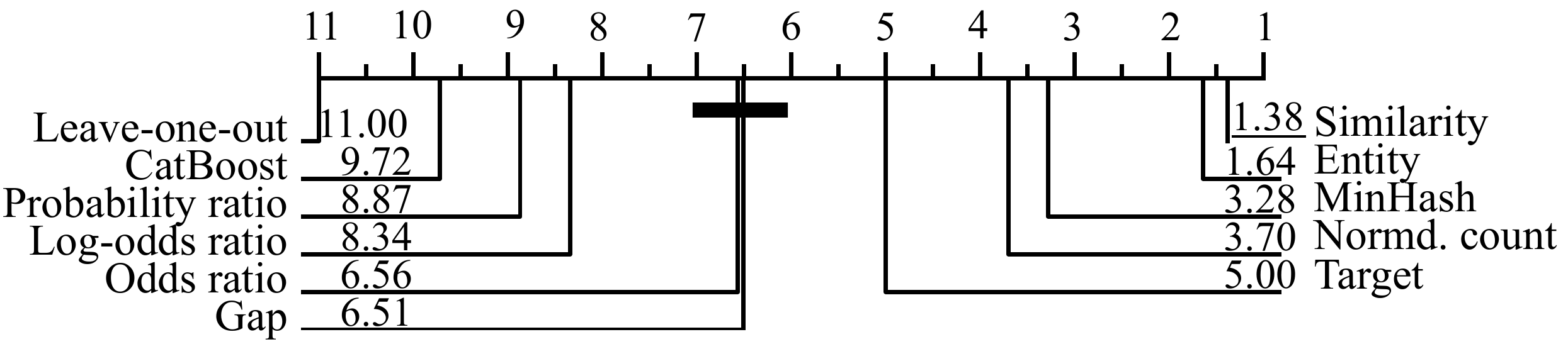}}
    \qquad
    \subfigure[\small Selecting the best model.]{\label{fig:cd_model_sel}%
      \includegraphics[scale=0.42]{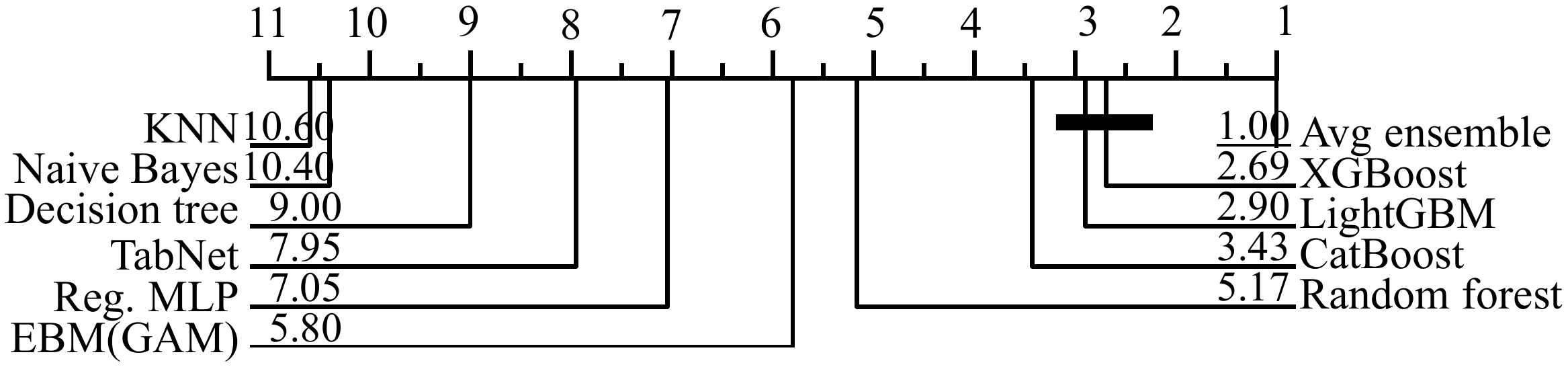}}
  }
\end{figure*}

We use a Critical Difference (CD) diagram~\citep{demvsar2006statistical} to represent the best encoder $E^*$ and model $f^*$ with the modeling split of the data (Figure~\ref{fig:cd_all}). The CD diagram is typically used to compare multiple models over multiple data sets statistically. To plot this, we  collect $1000$ bootstrap samples from the test set $\mathbf{X}^{\text{(ms,test)}}$, each of the same size as the test set. Then we run different encoders or models on each of these bootstrap samples yielding $1000$ values for AvRecall(10,40). The Friedman test is first performed to reject the null hypothesis (all models are statistically equivalent), followed by a post-hoc analysis based on the Wilcoxon significance test, in which critical differences are calculated. Further details on bootstrapping are given in Appendix~\ref{subsec:uncertainty}.


\subsection{Results on modeling split}
Most ($\sim 98\%$) features are categorical in nature. \autoref{fig:cd_all} and \tableref{tab:encodings} illustrate that for voluminous tabular data with mainly categorical features, some of which are of high cardinality, performance is quite sensitive to the encoding used. We find that similarity encoding performs the best among the extensive list of encoders we considered.
Figure~\ref{fig:cd_model_sel} shows a comparison of results of all models in terms of their \text{AvRecall(10,40)} performance using similarity encoding. Table~\ref{tab:models} lists appropriate metrics for these models; we find that 
the XGBoost model outperforms other individual models, with LightGBM and CatBoost being very close. 
Despite the recently reported encouraging performance of deep learning models on tabular data~\citep{TABNET,COCKTAILS}, we note that GBDT models still robustly outperform them on our large real-world tabular dataset (as noted in \citep{DLISNOT} as well). 

Note that, while Table~\ref{tab:models} displays the lift in ML model performance over the best baseline, the lift over random selection of patients for targeting is much more spectacular. For a random classifier, {\normalfont Recall}@k is $k/100$ and {\normalfont AvRecall(10,40)} is $0.25$. Hence, for the best model in Table~\ref{tab:models} (the average ensemble model), the percentage lift is $\sim 214\%$ ({\normalfont Recall}@20) and $\sim 172\%$ ({\normalfont AvRecall}(10,40)) from a random classifier. This is especially relevant because TB field staff currently are not known to follow specific standardized rules for patient prioritization.

We also find, as shown in Table~\ref{tab:models}, that the average ensemble of our top-5 best models slightly improves metrics over XGBoost. This may be attributed to the relatively low output correlation between EBM and the other models (Appendix \ref{app:subsec:ensembling}). On the other hand, we find that the solution is also quite robust in the sense of low predictive multiplicity across models  (Appendix \ref{app:subsec:pred_mult}). See  Appendix \ref{app:subsec:hyperparam_details} for further modeling details.

\begin{table*}[ht]
\centering
\floatconts
  {tab:encodings}%
  {\caption{\small Performance of different encoding techniques for categorical data using the XGBoost classifier. ``Lift'' denotes percentage increase in Recall@20 over the best baseline.}
    \vspace*{-2 em}
  }%
{
    \footnotesize
\begin{tabular}{cccc}
\toprule
\textbf{Encoding Type}           & \textbf{AvRecall(10,40)}          & \textbf{Recall@20}                    & \textbf{Lift (\%)} \\
\midrule
Leave-One-Out     & 0.373 ± 0.020  &	0.322 ±	0.022  &	6.05\%      \\
Probability Ratio & 0.535 ± 0.020 &	0.491 ± 0.023 &	56.37\%	      \\
Log-Odds Ratio  &  0.538 ± 0.021 & 0.502 ± 0.022 &	59.87\%	       \\
CatBoost Encoder & 0.523 ± 0.022 & 0.494 ± 0.024 & 57.32\%          \\
Odds Ratio  &  0.559 ± 0.021 & 0.524 ± 0.023	& 66.88\%             \\
Target Encoder & 0.603 ± 0.020 & 0.558 ± 0.023 & 77.71\%            \\
Normalized Count & 0.653 ± 0.020 & 0.614 ± 0.024 & 95.54\%           \\
Entity Embedding  &  0.670 ± 0.020 & 0.623 ± 0.025 & 98.41\%           \\
\textbf{Similarity Encoder} & \textbf{0.671 ± 0.019} & \textbf{0.624 ± 0.023} & \textbf{98.73\%}	\\
Gap Encoder    & 0.561 ± 0.021 & 0.508 ± 0.024 & 61.78\%                  \\
MinHash Encoder & 0.657 ± 0.020 & 0.614 ± 0.024 & 95.54\%   \\
\bottomrule
\end{tabular}%
}
\vspace{-1 em}
\end{table*}

\begin{table*}[ht]
\centering
\floatconts
  {tab:models}%
  {
  \vspace{-1.5 em}
  \caption{\small Performance of various ML models on the test set of the modeling split with similarity encoding with $95\%$ confidence intervals  generated from bootstrap samples. ``Lift'' is percentage increase in Recall@20 over best baseline.}
  \vspace{-2 em}
  }%
  {
  {%
     \footnotesize
\begin{tabular}{c c c c}
\toprule
{Model}           & {AvRecall(10,40)}          & {Recall@20}                    & {Lift(\%)} \\
\midrule
Random Forest       & 0.658 ± 0.019     & 0.613 ± 0.023 & 95.22 \\
$k$-NN                 & 0.481 ± 0.021     & 0.425 ± 0.024 & 35.35 \\
Decision Tree       & 0.515 ± 0.018     & 0.469 ± 0.023 & 49.36 \\
Naive Bayes         & 0.481 ± 0.018     & 0.408 ± 0.022 & 29.94 \\
\textbf{XGBoost}  & \textbf{0.671 ± 0.019}     & \textbf{0.624 ± 0.023} & 
    \textbf{98.73} \\
CatBoost & 0.668 ± 0.020     & 0.625 ± 0.024 & 99.04 \\
LightGBM            & 0.670 ± 0.020     & 0.620 ± 0.023 & 97.45 \\
Regularized MLP     & 0.612 ± 0.021     & 0.566 ± 0.024 & 80.25  \\
TabNet              & 0.602 ± 0.02      & 0.564 ± 0.023 & 79.62 \\
EBM (GAM)           & 0.650 ± 0.020     & 0.606 ± 0.025 & 92.99 \\
\textbf{Avg.\ Ensemble}  & \textbf{0.681 ± 0.019}     & \textbf{0.627 ± 0.025} & 
    \textbf{99.68}  \\
\bottomrule
\end{tabular}%
}
\vspace{-1 em}
}
\end{table*}

\subsection{Results on passive evaluation split}
\label{sec:evaluation}

We apply the best individual and ensemble trained models selected in the previous step to the passive evaluation holdout set $(\mathbf{y}^{(\text{pes}))}, \mathbf{X}^{(\text{pes})})$. We find that the XGBoost model applied to this set yields a Recall@20 of \textbf{0.614} and AvRecall(10,40) of \textbf{0.658}, while  the average ensemble model yields Recall@20 of \textbf{0.618} and AvRecall(10,40) of \textbf{0.663}. 
At $20\%$ patient targeting, assuming that all patients identified as at risk by the model are subjected to intensive interventions by health workers, this corresponds to saving $5587$ patients from becoming LFU over a six-month period across four Indian states, instead of $1808$ and $2974$ patients via the random classifier and best rule-based baseline yielding a lift of $\sim 209\%$ and $\sim 88\%$ respectively.

Table~\ref{tab:pe_cohort_month} displays the passive evaluation results broken down by month, signifying the generalization capabilities of the model trained on earlier data. There is no significant reduction in performance by month. Generalization capabilities across datasets spanning longer time periods are detailed in Appendix \ref{app:subsec:generalization}. Furthermore, Table~\ref{tab:pe_cohort_month} shows that the drop in performance on the passive evaluation time split relative to the modeling split test set (which was used to optimize model selection) (Table~\ref{tab:models}) is not appreciable.

We also carried out a detailed evaluation study of our best ensemble model across different data cohorts. For this, we use a global score threshold identified from the top-20\% of patients in the entire passive evaluation dataset, and an alternative to this global threshold by implementing a local district-level threshold (Appendix \ref{app:sec:cohort-wise-evaluation}).

\begin{table}[htb!]
\vspace{-1 em}
\centering
\floatconts
  {tab:pe_cohort_month}%
  {\caption{\small Passive evaluation results for ``Month of TB Treatment Initiation'' for $2020$.}
  \vspace{-1.5 em}
  }%
  {
  \footnotesize
\begin{tabular}{cccc}
\toprule

{Month} & {\#Patients} & {LFU Rate} &{Recall@20} \\
\midrule
Jul          & {48,208}              & 3.18\%        & {0.634 ± 0.024}      \\
{Aug}      & {42,797}  & {3.02\%} &  {0.604 ± 0.025}      \\
{Sep}      & {49,759}              & {3.05\%} &  {0.629 ± 0.024}      \\
{Oct}        & {51,639}              & {2.82\%} &{0.617 ± 0.025}      \\
{Nov}       & {50,288}              & {2.80\%}   & {0.614 ± 0.014}      \\
{Dec}       & {62,128}              & {2.51\%}  & {0.605 ± 0.025}     \\
\bottomrule
\end{tabular}%
}
  \vspace{-0.5 em}
\end{table}

We consider two types of cohorts in our detailed evaluation.

\noindent\textit{Geographical cohorts}: We analyze performance at the state and district levels. Beyond measuring the robustness of the model to geographical variations, this information is particularly useful in selecting geographies for deploying a pilot based on the model.

\noindent\textit{Non-geographical cohorts}: These cohorts are selected based on personal features, such as gender; temporal variables such as the month of treatment initiation and the time elapsed till the final outcome; and TB program-relevant cohorts such as type of healthcare system (public or private care) and type of the disease (drug-sensitive vs drug-resistant TB). We observe that our model performs exceptionally well on drug-resistant patients (the sub-group with the highest mortality rate) with a Recall@20 of \textbf{0.924} (\tableref{tab:pe_cohort_case}).

\begin{figure}[htb!]
\floatconts
  {image:recall-k}
  {
  \vspace{-1.5 em}
  \caption{\small $\text{Recall}@k$ vs $k$ for our best models, best baseline, and random classifier on the passive evaluation split.
  \vspace{-1 em}
  }}
  {\includegraphics[width=\linewidth]{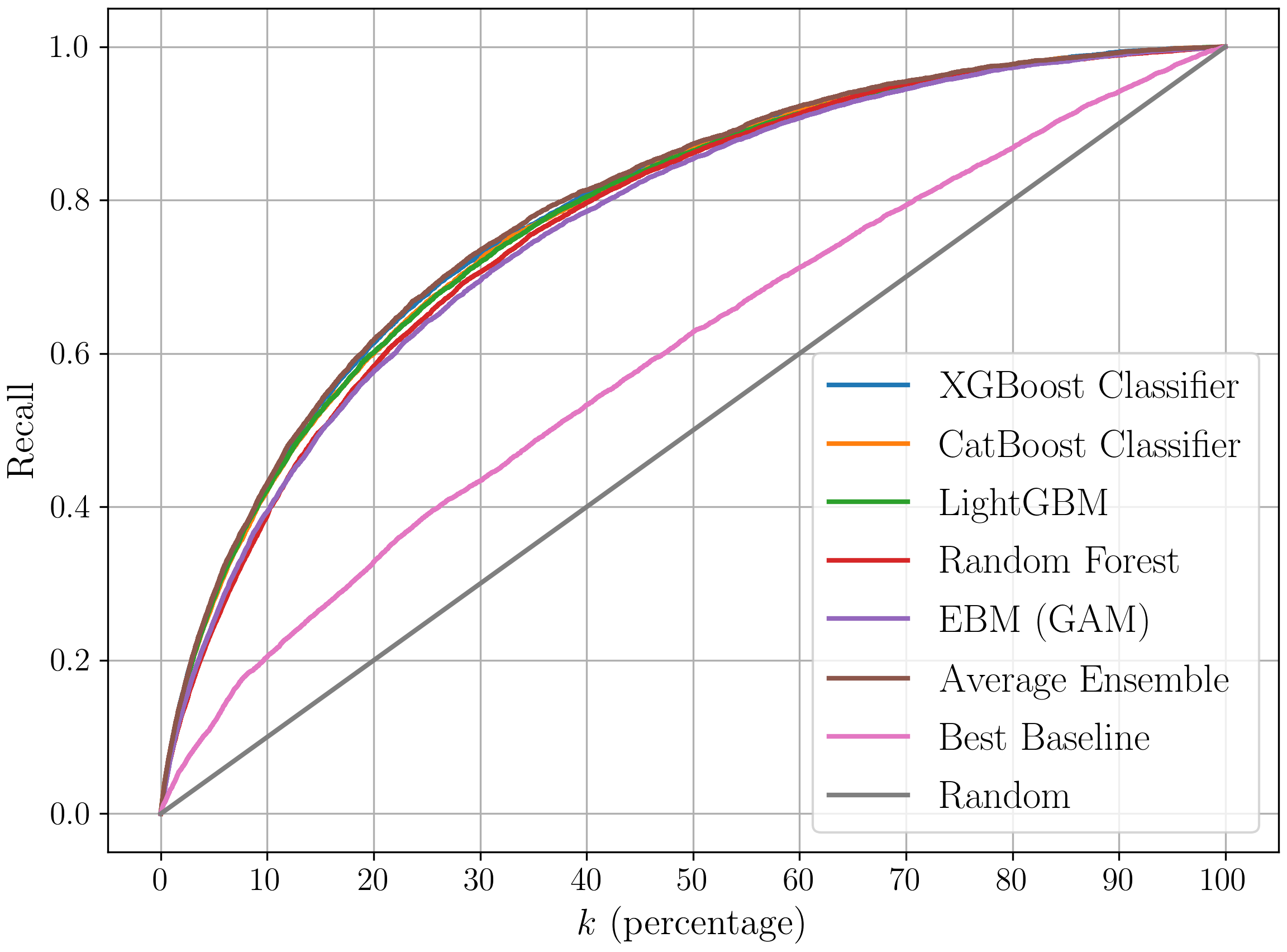}}
\vspace{-1 em}
\label{image:recall-k}
\end{figure}

While the value $k=20$ for $\text{Recall@k}$ and the range $k\in(10,40)$ for AvRecall are anecdotal, Figure~\ref{image:recall-k} shows that our models, although trained to optimize over AvRecall(10,40), perform well for all values of $k$.

The full results of cohort-based evaluation in Appendix \ref{app:sec:cohort-wise-evaluation} show that model performance is robust with respect to most cohorts of interest, except geographical cohorts and, to a lesser extent, gender. We discuss means of mitigating low performance in underperforming cohorts in \autoref{sec:improvements-underperforming-cohorts} below.

\noindent\textit{Other relevant metrics}: We tabulate results for other metrics for the best model and baseline in \tableref{tab:pe_other_metrics} for a more holistic view. The best model achieves satisfactory Precision@20 and AUC-PR (average precision) values for patient triaging with low target prevalence, significantly outperforming the baseline. A signal of good model performance is its high AUC-ROC, measuring quality of model score rankings across different thresholds. Here, the best model performs $33.1\%$ better than the baseline.

\begin{table}[htb!]
\vspace{-1 em}
\centering
\floatconts
  {tab:pe_other_metrics}%
  {\caption{\small Other relevant metrics on the passive evaluation set.}
  \vspace{-1.5 em}
  }%
  {
  \footnotesize
\begin{tabular}{cccc}
\toprule

{Metric} & {Best Model} & {Baseline} & {Gain} \\
\midrule
{Precision@20} & {0.090}  & {0.048}   & {87.5\%}      \\
{AUC-PR} & {0.134}  & {0.044}   & {204.5\%}      \\
{AUC-ROC} & {0.797}  & {0.599}   & {33.1\%}      \\
\bottomrule
\end{tabular}%
}
  \vspace{-1 em}
\end{table}

\subsection{Interpretability}

Interpretability is an important consideration in ML solutions for healthcare, both in terms of accountability and stakeholder involvement. For our use case, understanding important features is critical to designing effective interventions. To this end, 
 we implement Permutation Feature Importance (PFI)~\citep{PFI}, a global interpretability method, with respect to the Recall@20 metric (\tableref{tab:importances}, with details in Appendix \ref{app:subsec:interpretability_results}). The importance of the covariate \textit{BankDetailsAdded} for model performance is interesting.
This feature is in fact a proxy for patients on recurring monthly government monetary benefit schemes throughout their treatment course. This continuous positive reinforcement decreases LFU probability.

\begin{table}[ht]
\label{tab:perm_feat_imp}
\centering
\floatconts
  {tab:importances}%
  {
  \vspace{-1 em}
  \caption{\small Top-10 feature importance values with respect to Recall@20
using Permutation Feature Importance (PFI). More details in Appendix \ref{sub_sec:A_PFI}.}
\vspace{-1.5 em}
}%
  {
  \footnotesize
\begin{tabular}{ccc}
\toprule
{Feature}                      & {Importance}       & {Std.\ Dev.} \\
\midrule
BankDetailsAdded  & 0.105 & 0.002                      \\
ReasonforTesting&  0.060 & 0.002                      \\
CurrentFacilityPHIType                    & 0.029                         & 0.001                      \\
CurrentFacilityTBU                         & 0.026                         & 0.001                      \\
CurrentFacilityDistrict                               & 0.017                         & 0.001                      \\
CurrentFacilityPHI                                        & 0.015                        & 0.002                      \\
resultSampleA  & 0.012                        & 0.001                   \\
Age                      & 0.009                        & 0.002                       \\
TypeOfCase                                 & 0.008                          & 0.001                      \\
ContactTracing\_Done                       & 0.008                         & 0.001                     \\
\bottomrule
\end{tabular}%
\vspace{ -1 em}
}
\end{table}
We also implement the Local Interpretable Model-agnostic Explanations (LIME) \citep{LIME} method with a linear regressor for patient-level explanations, which also identifies BankDetailsAdded as a majority top feature. 
Appendix~\ref{app:subsec:interpretability_results} has further details and examples.

\subsection{Improving performance on underperforming cohorts}
\label{sec:improvements-underperforming-cohorts}

The model does reasonably well across most cohorts (see \autoref{app:sec:cohort-wise-evaluation}), but has variation across geographies and gender. We explore two approaches towards improving performance on underperforming cohorts and reducing variance; these are critical for our planned country-wide solution deployment.\\
\noindent $\bullet~\textit{Data augmentation}$: Augmenting the training data with copies of data from an underperforming cohort, similar to increasing the effective weight of patients from that cohort in the loss function. We do this at the district level, as discussed in Appendix~\ref{app:subsec:data-augmentation}.\\
\noindent $\bullet~\textit{Algorithmic fairness}$: Algorithmic fairness is a critical consideration in AI for healthcare settings.  We investigate a post-hoc fairness approach~\citep{rodolfa2021empirical} that adjusts model scores to equalize our metric across important cohorts such as states, districts, and gender, as detailed in Appendix~\ref{app:subsec:algorithmic-fairness}.

Both approaches significantly improve underperforming cohorts, especially the very low-scoring ones, with little impact on better ones, and a reduction in the overall variance. For instance, districts with Recall@20 less than $0.3$ show an $87\%$ lift in mean Recall@20 with augmentation, and $132\%$ with algorithmic fairness, albeit with an increase in effective $k$ for the latter (computed from \autoref{tab:data_aug} and \autoref{tab:algo_fair_underperforming_districts}). For gender, the fairness method also gives a modest improvement on the ``Female'' cohort (\autoref{tab:algo-fair-gender}).

\section{Responsible Deployment}
\label{sec:responsible-deployment}
We are currently planning pilots in multiple cities and states, with country-wide deployment to follow. We summarize some steps taken to ensure responsible deployment and leave a detailed description for future work.\\
\noindent $\bullet~\textit{Interventions}$: We propose risk-score-based interventions as per the draft differentiated care guidelines of India's Central TB division, with the consensus of on-ground stakeholders: local staff, and district and state-level TB officers.\\
\noindent $\bullet~\textit{Patient Safety}$: All patients are provided with standard care, with high-risk patients given extra care. We suggest not denying care to patients otherwise receiving it (via current interventions, e.g., counseling) and we ask staff to use their judgment as necessary. We clearly communicate that our tool is meant to improve, not replace, current practices. We also plan qualitative patient interviews and ground-truthing to verify outcomes.

\section{Conclusions}
\label{sec:key-takeaways}
 
This work describes the development of an ML solution for the prediction of LFU in TB patients. India, with the world's largest TB burden, and possessing one of the world's largest centrally administered longitudinal databases for tracking TB patients, represents a natural ground for testing and deployment of such solutions. The models were developed considering the complex deployment scenario, with data issues, distributional drift, and high cardinality features. We show that an ensemble combining the five best models is the best-performing across model classes (verified statistically) and important cohorts, and generalizes well across time. We explicitly address interpretability concerns and performance improvement using data augmentation and algorithmic fairness and find that the solution works especially well on DRTB patients, one of the most vulnerable patient cohorts. We also note some measures planned for ensuring the solution is deployed responsibly, to ensure patient benefit and safety.

Future technical directions include end-to-end learnable embeddings and fully interpretable modeling such as Neural Basis Models (NBM)~\citep{filipNBM}.  We are looking at extending to other adverse outcomes such as pre-treatment loss to follow-up, treatment failure, and mortality, potentially in a multiclass setting. Beyond classification, predicting time-to-LFU is a richer, more useful, but harder problem to solve. 

We hope that the results and insights from this work can be helpful to others working on similar problems and that its deployment can mark a successful milestone in the long journey toward tuberculosis eradication.

\section{Acknowledgment}

This work is made possible by the generous support of the American people through the United States Agency for International Development (USAID). The contents are the responsibility of Wadhwani AI and do not necessarily reflect the views of USAID or the United States Government. We are grateful to the Central TB division for providing us access to their data and for their continued support throughout the project. We also thank Harsh Raj, Devanshu Shah, Pranav Balaji, Mukul Kumar, Arpita Biswas, Makarand Tapaswi, Mayukh Nandy, and Amrita Mahale for their contributions. Finally, we thank all members, past and current, of the TB adherence team at Wadhwani AI for all their help.

\clearpage

\bibliography{kulkarni22}
\clearpage

\appendix

\section{Data Quality and Preprocessing}
\label{app:subsec:nikreg}

A brief description of the registers available in the Nikshay database is given in \autoref{tab:registers}, and the percentage of missing values in the data for each state is given in \autoref{tab:missing_values}.

\begin{table*}[h!]
\floatconts
  {tab:registers}%
  {\caption{\small Registers used to store patient-level data in Nikshay and the number of features from the registers used by the models. Features with low prevalence and potential leakage were removed during the pre-processing phase.}
  \vspace{-1 em}
  }%
  {\footnotesize
  \centering
\begin{tabular}{ m{2.2cm} m{10cm} m{1.5cm} }
\toprule
\bfseries Register & \bfseries Meaning & \bfseries \#Features \\
\midrule
Notification      & Basic details about the patient collected at notification time such as age, gender, location, health facility type, TB type, etc.                                                              & 54                                                              \\
Comorbidity       & Comorbidity information such as HIV and diabetes status, tobacco and alcohol addiction, etc.                                        & 17                                                              \\
Patient Data      & Demographic details such as occupation, marital status, etc.                    & 3                                                              \\
Adherence         & Information about adherence technology used & 1                                                              \\
Patient Lab       & Lab results such as CBNAAT, microscopy, culture, chest X-ray.                                               & 1                                                              \\
DMC               & Results from Designated Microscopy Centre (DMC)                                                                  & 3 \\ 
\bottomrule                                                           
\end{tabular}%
}
\end{table*}

\begin{table}[htbp]
\floatconts
  {tab:missing_values}%
  {
  \vspace{-1.5 em}
  \caption{\small State-wise percentage of missing/null values for all covariates across all patients in 2020.} 
  \vspace{-2 em}
  }%
  {\footnotesize
  \begin{tabular}{ccc}
  \toprule
  \bfseries State & \bfseries \#Patients & \bfseries Null Values\\
  \midrule
  Karnataka & 65,120 & 6.83\% \\
  Uttar Pradesh & 3,76,028 & 7.62\% \\
  West Bengal & 79,807 & 10.71\% \\
  Maharashtra & 1,57,997 & 6.85\% \\
  \midrule
  Total & 6,78,952 & 9.17\% \\
  \bottomrule
  \end{tabular}}
  \vspace{-1 em}
\end{table}

We used several strategies to select/remove columns, such as the trustworthiness and relevance of data, and the fraction of missing values. Most features used have less than $15\%$ missing values. However, we used some low-prevalence features identified to be important programmatically by public health domain experts. On the other hand, certain features had too many missing, noisy, or non-standardized values, perhaps owing to the difficulty in ensuring quality during collection in the field, and were excluded. For example, we exclude weight, a potentially important indicator, due to low prevalence. We also exclude certain features considered completely irrelevant to LFU. A number of features in the data are categorical, with some of them having  high cardinality. Thus, they are not fit for direct modeling, which motivates us to experiment with a large number of encoding techniques. Also, there are a number of data quality issues such as missing values, misspelled fields, etc., in several columns. We clean and transform the data in multiple steps. After removing features with a high proportion of missing or unintelligible string values, we use basic imputation techniques (mean imputation, creating an additional ``missing'' category) to fill in missing and invalid values in the remaining features. Finally, we merge all the registers based on a hashed $\text{EpisodeID}$ key (which acts as the primary key since a TB patient can experience multiple episodes of TB). We also consult domain experts to ensure that the categories have consistently named strings across registers and that there is no retrospective data in the input that can cause leakage.

\section{Rule-based Baselines}
\label{app:subsec:rule-based-baselines}
\begin{table*}[h!]
\floatconts
    {tab:baselines}%
  {\caption{\small Rule-based baseline models, and their performance on the test set of the modeling split.}
  \vspace{-1 em}
  }%
  {%
  \footnotesize
    \centering
    \begin{tabular}{m{2.2cm} m{10cm} m{1.8cm}}
    \toprule
    \bfseries Baseline & \bfseries Features & \bfseries Recall@20\\
    \midrule
    Rule 1 &  
    Alcohol intake history (present),
    HIV status (positive),
    Diabetes status (positive) & 0.184 \\
    \midrule
    Rule 2 &
    Type of Case (DRTB or retreatment),
    HIV status (positive),
    Diabetes Status (positive),
    Age ($\geq 60$),
    Household contacts above age six (absent) & 0.282 \\
    \midrule
    Rule 3 &
    Age ($\in [18,45])$, 
    Type of Case (DRTB or retreatment), 
    HIV Status (positive), 
    Diabetes Status (positive), 
    Gender (male or transgender), 
    Migrant:change of district after diagnosis (yes) , 
    Alcohol intake history (present), 
    Household contacts above age six (absent), 
    Microbiologically Confirmed (yes), 
    Disease site (pulmonary), 
    UDST Done (yes), 
    Bank Details Added (no) & 0.314 \\
    \bottomrule
    \end{tabular}
    }
    \vspace{-1.5 em}
\end{table*}

Table~\ref{tab:baselines} lists features included in the three rule-based baseline models. The baseline scores are defined to be the fraction of features that lie within a specified range. Rule 1 comes from NTEP draft guidelines (unpublished), Rule 2 from guidelines proposed by \citep{khpt}, while the best-performing baseline (Rule 3) comes from an extensive survey of the literature~\citep{jaggarajamma2007reasons, bhatnagar2019user, bhattacharya2018barriers}, and analysis of features that correlate the most with the LFU outcome in the Nikshay data. 

\section{Metric Details}
\label{sec:metric_discussion}
\subsection{Recall@k}

We note that $\text{Recall@k}$ has an upper bound of $\min(1,k/p)$, where $p$ is the prevalence of the positive class. This bound, however, is not of concern for us since the prevalence is considerably less than the percentage targeted. A model that assigns random risk scores from $[0, 1]$ to all patients is expected to have a $\text{Recall@20}$ of $0.20$.

\subsection{Ambiguity and discrepancy}
\label{app:subsec:pred_mult}

We see in \tableref{tab:models} that a number of models perform similarly on our aggregated, rank-based metric. However, on individual patients, it is possible that these models give very different scores. Predictive multiplicity is defined as the ability of a prediction problem to admit competing models with conflicting predictions. For evaluating the robustness of our solution against model selection and improving stakeholder participation, we compute two metrics of predictive multiplicity \citep{PRED_MULT} for the set of our best-performing models. Let $h_0$ be the best performing model, with performance metric $p$, and $S_\epsilon(h_0)$ be a set of models with a performance range of $[(1 - \epsilon)p, p]$.  we define Ambiguity ($\alpha_\epsilon$) and Discrepancy ($\delta_\epsilon$) as follows:
\begin{gather*}
\alpha_\epsilon(S_\epsilon) \mathrel{:\mkern-0.25mu=} \frac{1}{n} \sum_{i=1}^{n} \max_{h \in S_\epsilon(h_0)} \mathbbm{1} [h(\mathbf{x_i}) \neq h_0(\mathbf{x_i})],\\
\delta_\epsilon(S_\epsilon) \mathrel{:\mkern-0.25mu=} \frac{1}{n} \max_{h \in S_\epsilon(h_0)} \sum_{i=1}^{n} \mathbbm{1} [h(\mathbf{x_i}) \neq h_0(\mathbf{x_i})],
\end{gather*}

\noindent where $n$ is the number of patients. We take an $\epsilon = 0.2$ set of similarly performing models, and find that they have an ambiguity score of $22.24\%$ and a discrepancy score of $13.51\%$ on the test set of the modeling split. Comparing with the same measures on other real-world datasets from \citep{PRED_MULT}, we find these values well within the acceptable range.
\section{Modeling Details}
\label{app:subsec:hyperparam_details}

\subsection{Hyperparameter search spaces}

We run a $100$-iteration Hyperopt hyperparameter search on every model. The hyperparameter search space used for XGBoost models is given in \tableref{tab:xgb_hparams}.

For the regularized MLP model, as suggested by \citep{COCKTAILS}, we create an MLP with $5$ fully-connected layers with a skip connection from the second layer to the last layer. Hyperparameters such as the size of the layers, the learning rate, the optimizer, the weight of the low-prevalence class in the loss function, dropout, SGD momentum, gradient clip, early stopping, learning-rate scheduling, etc. are optimized through TPE. Further details for all models are made available through a shared repository.\footnote{For the complete list of hyperparameter search spaces for all our models, please visit: \url{https://anonymous.4open.science/r/model_configurations-B003/}}

\begin{table*}[htbp]
\centering
\floatconts
  {tab:xgb_hparams}%
  {\caption{\small The hyperparameter search space for XGBoost models. Optimization was done using TPE on the AvRecall(10, 40) values on the validation set of the modeling split.}
  \vspace{-1 em}
  }%
  {
  \footnotesize
\begin{tabular}{ccc}
\toprule
\textbf{Hyperparameter}      & \textbf{Distribution} &\textbf{Range} \\ 
\midrule
Learning rate & Log Uniform & $[10^{-7}, 1]$ \\
Num. estimators & Discrete Uniform & $\{50, 100, 200, 400, 600, 800, 1000, 1200, 1500, 2000\}$ \\
Maximum depth & Uniform Integer  & $[1, 9]$ \\
Min child weight & Uniform Integer  & $[1, 8]$ \\
Scale pos. weight & Uniform Integer & $[1, 90]$ \\
L1 term & Log Uniform & $[10^{-5}, 1]$ \\
L2 term & Uniform & $[10^{-3}, 1]$ \\
Subsample ratio & Uniform & $[0.5, 1]$ \\
Col sample ratio & Uniform & $[0.5, 1]$ \\
Min split loss & Discrete Uniform & $\{0, 0.3, 0.6, 0.9, 1.2, 2, 4, 6, 8\}$ \\
\bottomrule
\end{tabular}
}
\end{table*}

\subsection{Ensembling}
\label{app:subsec:ensembling}

On analyzing model scores, we notice a large correlation between the GBDTs, which all have a relatively low correlation with EBM as shown in Figure~\ref{fig:Model_Corr}. To try and exploit this, we investigate ensembling EBM with XGBoost, Catboost, LightGBM, and Random Forest.  An averaged ensemble leads to a small boost in performance (\autoref{tab:models}).

\begin{figure}[ht]
\floatconts
  {fig:Model_Corr}
  {
  \vspace{-2 em}
  \caption{\small Correlation matrix for our top-5 models with similarity encoding on the passive evaluation split.}}
  {\includegraphics[width=\linewidth]{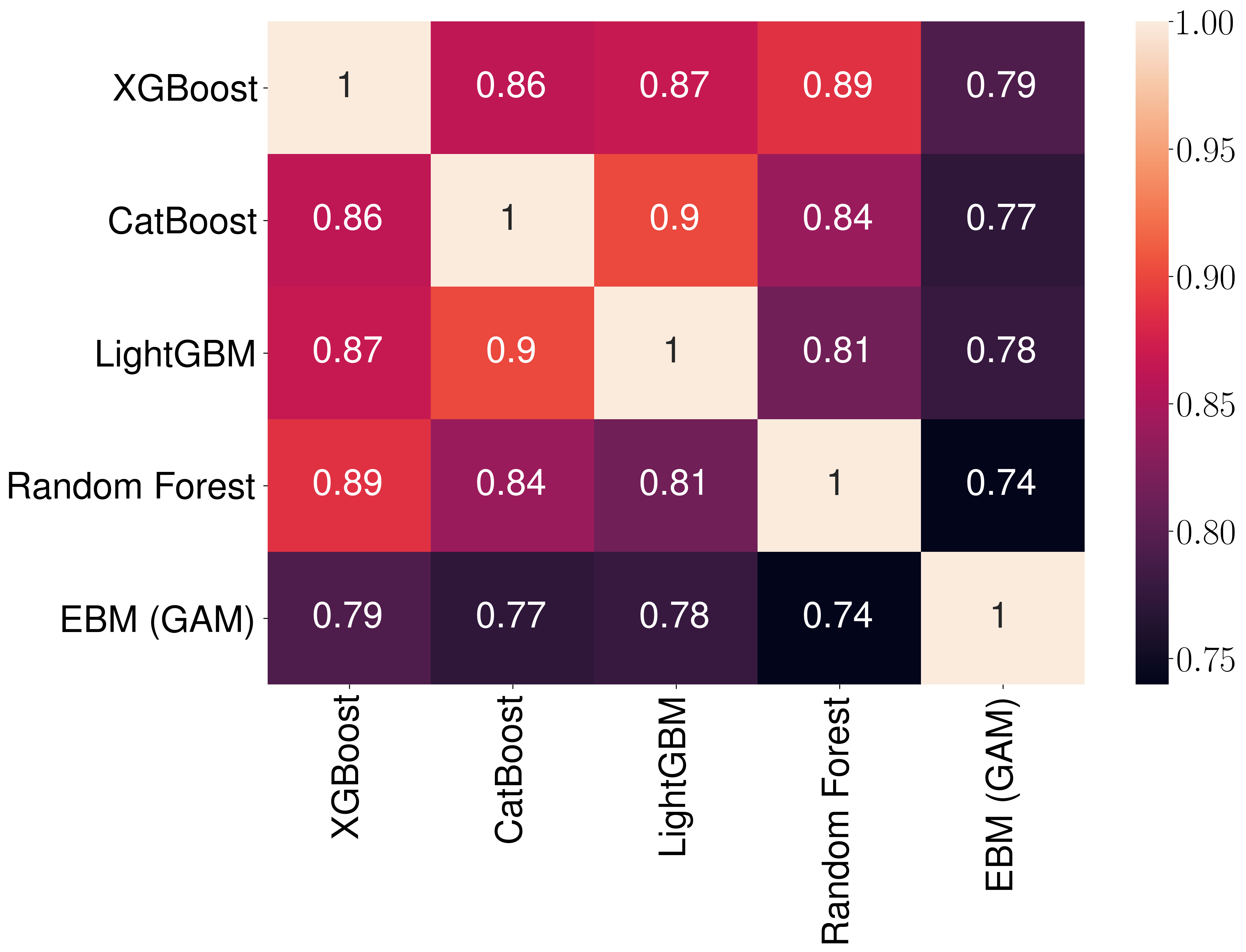}}
\vskip -0.2in
\end{figure}



\section{Temporal Generalization Analysis}
\label{app:subsec:generalization}
We carried out various experiments to test the ability of the model to generalize across time. In particular, given the COVID-19 pandemic that began in 2020 which had the potential to significantly affect patient behaviour, this analysis is important to understand how well the model would perform outside these abnormal circumstances.

First, we initially obtained a smaller dataset of patients from the year $2019$ in order to establish a proof of concept. We split this into a training set of $38,000$ patients and a holdout set of $3,800$ patients (this evaluation is termed PoC-2019). 
We later gained access to all of the $2019$ data and $2020$ data. We received access to four registers -- notification, comorbidity, contact tracing, and patient lab registers -- for the year 2019, while for 2020 data, three additional registers were shared. As described in Section~\ref{subsec:forward_splitting}, the data was split temporally in a forward fashion to mimic actual deployment. We performed passive evaluation on these datasets (termed PE-2019 and PE-2020). Results from these evaluations are summarized in Table~\ref{tab:evaluations}, and show good temporal generalization. 

\begin{table}[htbp!]
\floatconts
  {tab:evaluations}%
  {
  \vspace{-1 em}
  \caption{\small Summary of results on different datasets received. We note that Recall@20 is roughly constant across these evaluations. The superscript to $n$ indicates the data split.} 
  \vspace{-2 em}
  }%
  {\footnotesize
  \begin{tabular}{ccccc}
  \toprule
  \bfseries Evaluation & \bfseries $n^{\text{(ms)}}$ & \bfseries $n^{\text{(pes)}}$ & \bfseries Recall@20\\
  \midrule
  PoC-2019 & 38,243 & 3800 & 0.581 \\
  PE-2019 & 943,654 & 431,932 & 0.569 \\
  PE-2020 & 678,952 & 325,190 & 0.618 \\
  \bottomrule
  \end{tabular}}
  \vspace{-1 em}
\end{table}
We then designed the following experiments with our best model class from $2019$, using the four registers available from then:

\noindent \textbf{Experiment 1:} Best model trained with the first six months of the 2019 data and evaluated separately on the last six months of 2019 and the first six months of 2020.

\noindent \textbf{Experiment 2.1:} Model trained on all of the 2019 data and tested on the sixth month of the 2020 data. 

\noindent \textbf{Experiment 2.2:} Model trained on 2019 and the first five months of 2020 data and tested on the sixth month of 2020 data.

The aim of Experiment 1 is to understand how robust a previously trained model is on new data, and that of Experiments 2.1 and 2.2 is to understand the importance of the availability of immediately preceding data in the performance of the model. The result of Experiment 1 shows that the model is inherently robust and generalizes well across time as the Recall@20 value remained consistent from 2019 to 2020 (\tableref{tab:Gen_Expt_1}). On the other hand, the results of Experiments 2.1 and 2.2 suggest a significant increase in performance when the model is trained using the immediately preceding data (\tableref{tab:Gen_Expt_2}).

\begin{table}[ht]
\centering
\floatconts
  {tab:Gen_Expt_1}%
  {\caption{\small Comparison of Recall@20 in Experiment 1, showing the generalization ability of the model. Training on the first half of  2019 data leads to similar performance on the second half of 2019 data and the first half of 2020 data.}
  \vspace{-1 em}
  }%
  {\footnotesize
\begin{tabular}{ccc}
\toprule
\textbf{Experiment 1} & \textbf{2019} & \textbf{2020} \\
\midrule
Recall@20         & 0.560               & 0.578                                        \\
\bottomrule
\end{tabular}%
}
\end{table}

\begin{table}[ht]
\centering
\floatconts
  {tab:Gen_Expt_2}%
  {\caption{\small Comparison of Recall@20 between Experiments 2.1 and 2.2. For the same test set, a model that is trained on the immediately preceding data outperforms a model that is trained much earlier.}
  \vspace{-1 em}
  }%
  {\footnotesize
\begin{tabular}{cc}
\toprule
\textbf{Experiment 2} & \textbf{Test Set Recall@20} \\
\midrule
Experiment 2.1         & 0.572                                        \\
Experiment 2.2         & 0.606                                       \\
\bottomrule 
\end{tabular}%
}
\end{table}

We observe that the model is quite robust on later data and also that the addition of the latest data helps it adapt to distribution shifts.

\section{Cohort-Wise Evaluation of Model Performance}
\label{app:sec:cohort-wise-evaluation}

We classify patients into two categories based on the average ensemble model score: high-risk and low-risk, with the corresponding thresholds calculated in accordance with the rank-based metric Recall@20. The top $20$ percentile of the patients based on the model scores are classified as high-risk, with the remaining classified as low-risk. We explore two methods of thresholding:
\begin{itemize}
    \item \textit{Global}: This is a single threshold corresponding to the top-20 percentile of all the patients being evaluated, which is applied universally.
    \item \textit{Local}: We use a local district-level threshold to ensure that every district has a consistent 20\% of patients targeted. Thresholds thus calculated are per district and are applied to patients in those districts only.
\end{itemize}
We define effective $k$ as the percentage of patients targeted in a particular cohort, which would be the $k$ corresponding to using a cohort-level threshold. While a global threshold leads to higher overall sensitivity, it results in  different levels of targeting at different districts. This leads to insufficient targeting and poor sensitivity with low effective $k$ in some districts, and an impractically high effective $k$ in some others. We use local thresholding to fix these issues, and consider it more relevant for an on-ground deployment. We also consider using local thresholds at even more granular (sub-district) levels but do not proceed with it because of the small population sizes for these cohorts leading to a large variance in the predictions.

While we observe similar performance across both thresholding methods, we also analyze how they affect the performance in some major cohorts. A few cohorts of importance that we assess are:
\begin{itemize}
    \item The month of treatment initiation
    \item Gender
    \item Type of Case (drug-sensitive vs drug-resistant)
    \item Peripheral health institution (PHI) type (public vs private)
\end{itemize}
We observe that the effective $k$ (targeted \%) is roughly constant across the six different months of evaluation, as reported in \tableref{tab:pe_cohort_month_effective_k}, indicating robust performance across time when using a global threshold. However, we also observe that there is a large difference in effective $k$ across the 4 states as shown in \tableref{tab:pe_cohort_state}. This leads to a large difference in Recall@20 across these states. On the other hand, local thresholding enables the model to target every state equally and thus boosts the performance in under-performing states (\tableref{tab:pe_cohort_state}). We also study the model performance and implications of threshold type used on all non-location cohorts. 

For example, the performance of the model stays largely constant on the ``Gender'' cohort (\tableref{tab:pe_cohort_gender}).  We observe that there is an overall improvement in the performance of the model on the ``Male'' and ``Female'' cohorts if we use a local threshold. It does, however, lead to a trade-off with a drop in performance on the ``Transgender'' category. We observe a similar trade-off with the ``Type Of Case'' cohorts (\tableref{tab:pe_cohort_case}). We find that the model has a high sensitivity for LFU prediction on patients who are diagnosed with DRTB (Drug-resistant TB) as compared to DSTB (Drug-sensitive TB patients) in both cases. Using a local threshold results in an increase in performance across the DSTB patients, but that is traded-off by a decrease in performance across the DRTB patients. Since DRTB patients have a higher propensity to become LFU, and have a separate, longer treatment regimen with more intensive care, this is particularly relevant when choosing between the two thresholding methods. On the other hand, we don't observe such trade-offs in the ``PHI Type'' cohort. This cohort is a useful axis of measurement since the public and private systems are both important while having different ways of working. We observe that local thresholding outperforms global thresholding (\tableref{tab:pe_cohort_phi_type}) across both its categories.

While our model performs satisfactorily on all non-location cohorts, we observe that there are multiple trade-offs that need to be considered while choosing between the two thresholding methods. On-ground personnel can, at the district-level, implement further granular thresholding across non-location cohorts (Eg., ``Gender'') to either improve or equalize the performance across that cohort.

\begin{table*}[ht]
\centering
\floatconts
  {tab:pe_cohort_state}%
  {\caption{\small Passive evaluation results for the location cohort ``State'' with a global threshold vs a local district-level threshold.}
  \vspace{-1 em}
  }%
  {
  \footnotesize
\begin{tabular}{cccc}
\toprule
\textbf{State} & \textbf{Effective k (Global)} & \textbf{Recall@20 (Global)} & \textbf{Recall@20 (Local)} \\
\midrule
{Karnataka} & {11.981 ± 0.306} & {0.516 ± 0.020} & {0.668 ± 0.027}      \\
{Maharashtra} & {14.224 ± 0.192} & {0.607 ± 0.013} & {0.682 ± 0.013}      \\
{Utar Pradesh} & {25.387 ± 0.149} & {0.669 ± 0.006} & {0.614 ± 0.008}     \\
{West Bengal} & {10.954 ± 0.285} & {0.403 ± 0.019} & {0.567 ± 0.014}     \\
\bottomrule
\end{tabular}%
}
\end{table*}

\begin{table*}[ht]
\centering
\floatconts
  {tab:pe_cohort_gender}%
  {\caption{\small Passive evaluation results for the ``Gender'' cohort with a global threshold vs a local district-level threshold.}
  \vspace{-1 em}
  }%
  {
  \footnotesize
\begin{tabular}{ccccc}
\toprule
\textbf{Gender} & \textbf{\#Patients} & \textbf{\#LFUs} & \textbf{Recall@20 (Global)} & \textbf{Recall@20 (Local)} \\
\midrule
{ Female}          & { 130,068}                                                         & { 3,201}                                                       & {0.563 ± 0.008}       & {0.576 ± 0.014}      \\
{ Male}            & { 181,277}                                                         & { 5,837}                                                       & {0.645 ± 0.007}       & {0.658 ± 0.005}      \\
{ Transgender}     & { 137}                                                             & { 4}                                                           & {0.674 ± 0.326}       & {0.552 ± 0.219
}     \\
\bottomrule
\end{tabular}%
}
\end{table*}

\begin{table*}[ht]
\floatconts
  {tab:pe_cohort_case}%
  {\caption{\small Passive evaluation results for the ``Type of Case'' cohort with a global threshold vs  a local district-level threshold.}
  \vspace{-1 em}
  }%
  {\footnotesize
\begin{tabular}{ccccc}
\toprule
\textbf{Type Of Case} & \textbf{\#Patients} & \textbf{\#LFUs} & \textbf{Recall@20 (Global)} & \textbf{Recall@20 (Local)} \\
\midrule
Drug-resistant TB                & 10,267                                                          &                                             909           & 0.924 ± 0.011
      & 0.907 ± 0.013
      \\
Drug-sensitive TB & 301,215 & 8,133 & 0.586 ± 0.007       &  0.601 ± 0.008     \\
\bottomrule
\end{tabular}
}
\end{table*}

\begin{table*}[ht]
\floatconts
  {tab:pe_cohort_phi_type}%
  {\caption{
  \small Passive evaluation results for the ``PHI Type'' cohort with a global threshold vs  a local district-level threshold.}
  \vspace{-1 em}
  }%
{%
\footnotesize
\begin{tabular}{ccccc}
\toprule
\textbf{PHI Type} & \textbf{\#Patients} & \textbf{\#LFUs} & \textbf{Recall@20 (Global)} & \textbf{Recall@20 (Local)} \\
\midrule
Private           & 87,910 & 2,443                                                       & {0.598 ± 0.009}       & {0.610 ± 0.015}      \\
{ Public}            & { 223,572}                                                         & { 6,599}                                                       & {0.626 ± 0.008}       & {0.641 ± 0.006}     \\
\bottomrule
\end{tabular}%
}
\end{table*}

\begin{table}[htb!]
\vspace{-1.5 em}
\centering
\floatconts
  {tab:pe_cohort_month_effective_k}%
  {\caption{\small Passive evaluation results for the cohort ``Month of TB Treatment Initiation'', for the year $2020$, with effective $k$}
  \vspace{-2 em}
  }%
  {\footnotesize
\begin{tabular}{ccc}
\toprule

{Month} &{Effective k} &{Recall@20 (Global)} \\
\midrule
Jul  &  { 19.919 ± 0.358}    & {0.634 ± 0.024}      \\
{Aug}   &  {19.579 ± 0.359}  & {0.604 ± 0.025}      \\
{Sep}   & {20.091 ± 0.364}   & {0.629 ± 0.024}      \\
{Oct}      &{20.092 ± 0.347}  & {0.617 ± 0.025}      \\
{Nov}    &{20.976 ± 0.347}   & {0.614 ± 0.014}      \\
{Dec}       & {19.572 ± 0.319}  & {0.605 ± 0.025}     \\
\bottomrule
\end{tabular}%
}
\end{table}

\section{Interpretability Methods}
\label{app:subsec:interpretability_results}
\subsection{Permutation Feature Importance (PFI)}
\label{sub_sec:A_PFI}

\tableref{tab:importances_meaning} shows the top-10 features determined by the PFI method on the best model, along with their definitions.

\begin{table*}[h!]
\label{tab:perm_feat_imp_meaning}

\floatconts
  {tab:importances_meaning}%
  {
  \vspace{-1 em}
  \caption{\small Top-10 feature importance values with respect to Recall@20
using Permutation Feature Importance (PFI).}
\vspace{-2 em}
}%
  {
    \footnotesize
  \centering

\begin{tabular}{m{3.3cm} m{8cm} m{1.6 cm} m{1.25cm} }
\toprule
{ \textbf{Feature}}      & \textbf{{Meaning}}                  & \textbf{{Importance}}       & \textbf{{Std. Dev.}} \\
\midrule
BankDetailsAdded  & This is a binary ``yes/no'' feature representing whether patients have their bank details added to Nikshay, signifying an increased likelihood of receiving monthly monetary government incentives for better nutrition during the TB treatment.
& 0.105 & 0.002                      \\
\midrule
ReasonforTesting & Whether the test offered is for diagnosing TB or for assessing disease prognosis using follow-up testing. 
& 0.060 & 0.002                      \\
\midrule
CurrentFacilityPHIType  & Binary variable: public/private. ``Current'' refers to the patient’s preferred place of receiving medication and follow up. PHI (Peripheral Health Institute) is a health facility manned by at least one officer.  & 0.029                         & 0.001                      \\
\midrule
CurrentFacilityTBU &  TBU is an administrative unit under the program with standardized coverage area, population, and resources budgeted under the program.                         & 0.026                         & 0.001                      \\
\midrule
CurrentFacilityDistrict & The district where a patient is seeking TB treatment and care, which may or may not be the place from where they originate get diagnosed. & 0.017                         & 0.001                      \\
\midrule
CurrentFacilityPHI      & Refer ``CurrentFacilityPHIType''. This refers to the actual PHI accessed by the patient.                                & 0.015                        & 0.002                      \\
\midrule
resultSampleA  & The specimen used for testing TB is received in two samples. This refers to the result for the first sample. & 0.012                        & 0.001                   \\
\midrule
Age      &  Age of the patient. & 0.009                        & 0.002                       \\
\midrule
TypeOfCase   &  Drug sensitive TB vs Drug-resistant TB.                              & 0.008                          & 0.001                      \\
\midrule
ContactTracing\_Done   & Whether the treatment supervisor visits the patient’s home and assesses family contacts who are living in the same house. These contacts are at an increased risk of acquiring TB from them and are thus subject to mandatory screening for TB. & 0.008                         & 0.001                     \\
\bottomrule
\end{tabular}%
\vspace{ -1 em}
}
\end{table*}

\subsection{Accumulated Local Effects (ALE)}
\label{app:subsec:ALE}
We implement global interpretability techniques such as an Accumulated Local Effects (ALE) \citep{ALE} plot to visually determine the effect of each feature on the predicted target. The ALE Plot for the feature ``Age'' can be viewed in \figureref{fig:ALE_Age}. We observe that there is a steep increase in risk as the age increases from $18$ to $30$, followed by a similar performance till the age of $60$ and then a gradual increase in risk beyond age $60$. This is consistent with reports from domain experts and helpful in creating appropriate brackets for age-specific interventions.

\begin{figure}[ht]
\floatconts
  {fig:ALE_Age}
  {
  \vspace{-2 em}
  \caption{\small ALE Plot of the feature ``Age'' with respect to the model outcome.}}
  {\includegraphics[width=\linewidth]{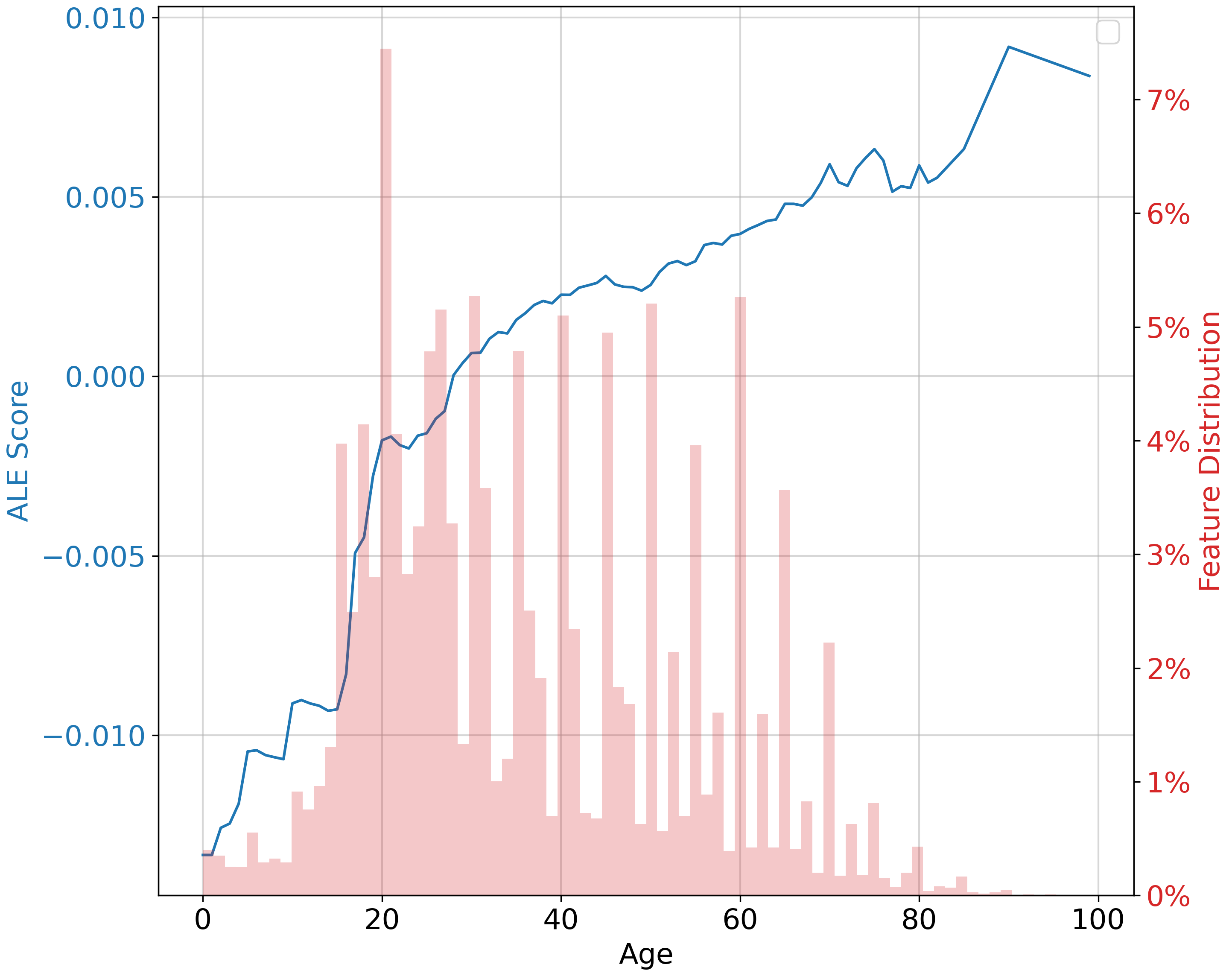}}
\vskip -0.2in
\end{figure}

\subsection{Local Interpretable Model-agnostic Explanation (LIME)}
\label{app:subsec:LIME}

The LIME result for a randomly selected patient can be viewed in  \figureref{fig:LIME_High}. The patient has a model score of 0.037 and a ground truth label of 0. We gain valuable insights into the model behavior by analyzing the LIME output. Features with positive weights (green) help the model to push the decision towards LFU, while those with negative weights (red) help push it towards non-LFU. We see that \textit{BankDetailsAdded}, which is a proxy for whether the patient is receiving direct cash transfers, plays a major role in reducing the score of the patient, and that several location features have an impact on the prediction.

\begin{figure}[ht]
\floatconts
  {fig:LIME_High}
  {
  \vspace{-1 em}
  \caption{\small Feature importance values for a randomly selected patient prediction generated using LIME with a Linear Regressor. The feature importances show the impact that various features to give an output of 0.037 for a non-LFU Patient.}}
  {\includegraphics[width=\linewidth]{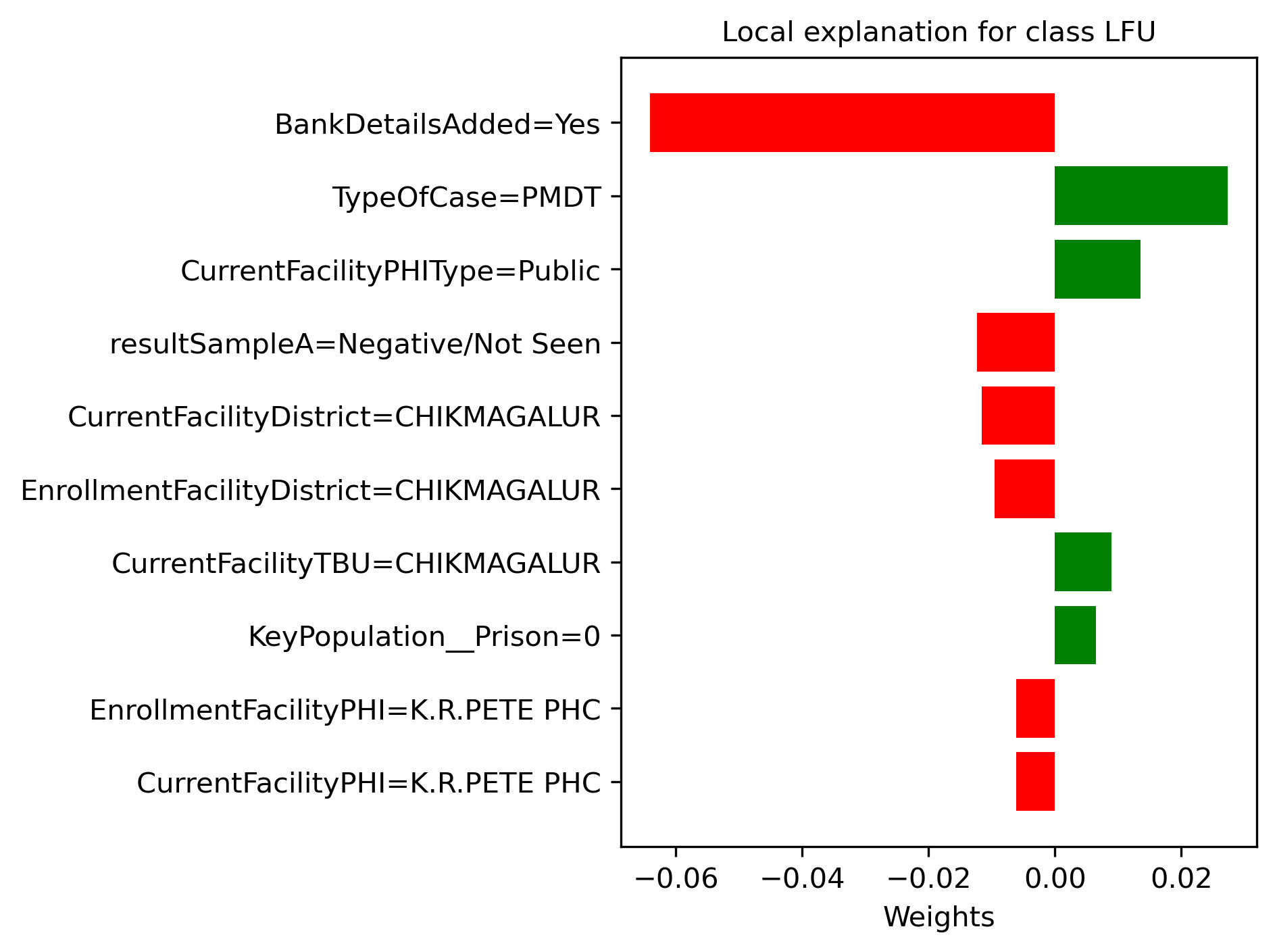}}
\vskip -0.2in
\end{figure}

\section{Uncertainty Calculation}
\label{app:sec:uncertainty-calculation}
We use the following algorithm to generate bootstrapped samples:

\begin{itemize}
    \item For a cohort $C$ of size $N$, draw 1000 $N$-sized samples, $s_1, \ldots , s_{1000}$ with replacement from $C$.
    \item Run the model on the $s_i$'s to generate 1000 $N$-sized lists of model scores $l_1, \ldots , l_{1000}$.
    \item Compute our metric of interest (e.g. Recall@20, $\text{AvRecall}(10,40)$) on the $l_i$'s, generating 1000 samples of our metric $m_1, \ldots , m_{1000}$.
\end{itemize}
The $m_i$'s can be used for generating uncertainty intervals, boxplots (see \figureref{fig:uncertainty-boxplots-models} for models and \figureref{fig:uncertainty-boxplots-encoders} for encoders), critical diagrams, etc. Note that the training set is fixed throughout this pipeline and we do not use bootstrapping during model selection (on the val set) either.
See \figureref{fig:uncertainty-interval-methods} for an example of generated uncertainty intervals using bootstrap, using confidence intervals obtained from the empirical distribution and a fitted Gaussian. We note that 95\% confidence intervals for both methods are nearly identical.
\label{subsec:uncertainty}

\begin{figure}[htb!]
\floatconts
  {fig:uncertainty-interval-methods}
  {
  \vspace{-2 em}
  \caption{\small Empirical distribution with best-fit Gaussian for AvRecall(10,40) with 1000 bootstrap samples, with 95\% uncertainty intervals. This is for the best model-encoding pair on the modeling split test set. }}
  {\includegraphics[width=\linewidth]{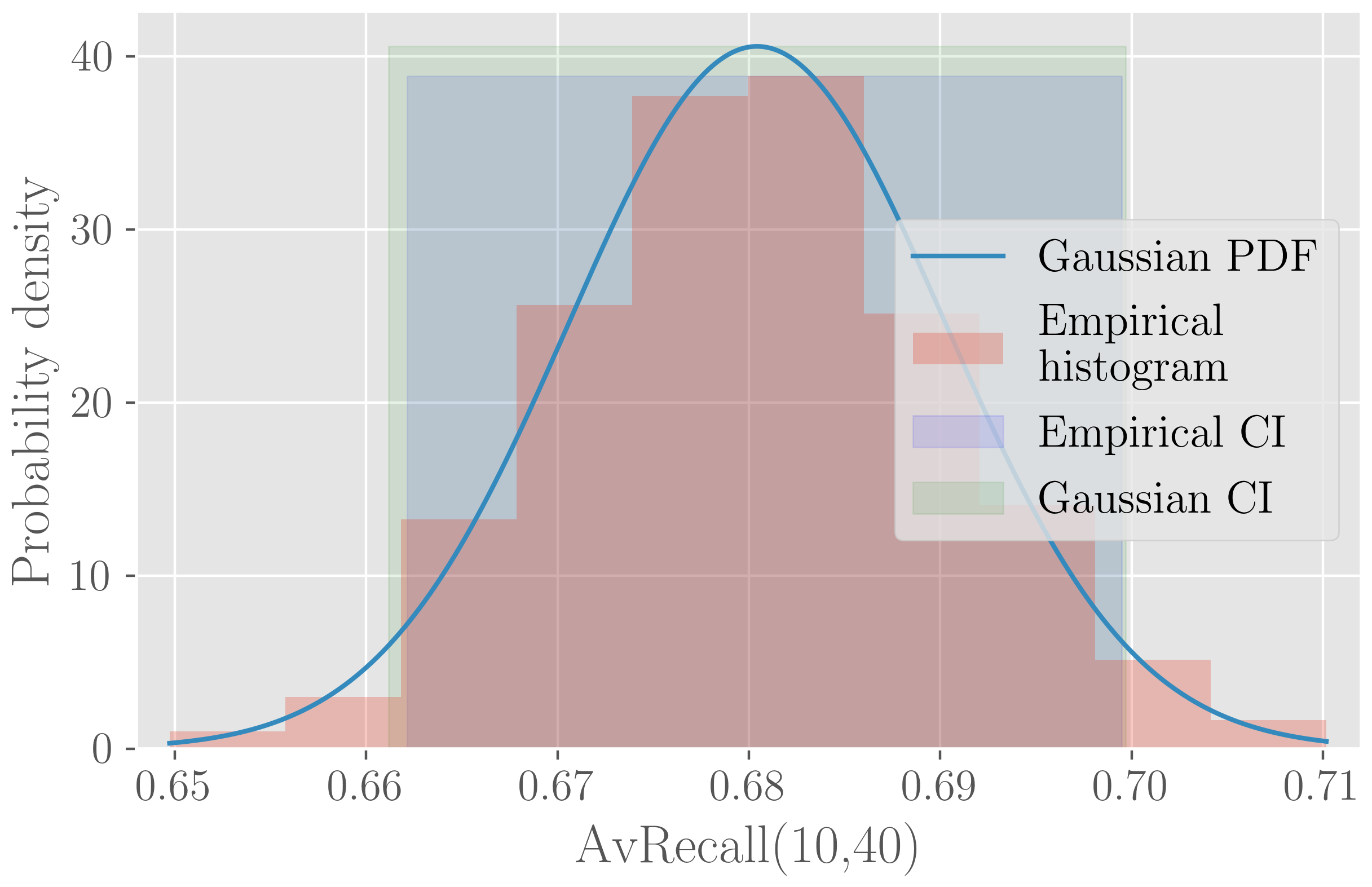}}
\vskip -0.1in
\end{figure}

\begin{figure}[htb!]
\floatconts
  {fig:uncertainty-boxplots-models}
  {
  \vspace{-2 em}
  \caption{\small Boxplots with 1000 bootstrap samples for various models used with the similarity encoder, generated from the modeling split test set.}}
  {\includegraphics[width=\linewidth]{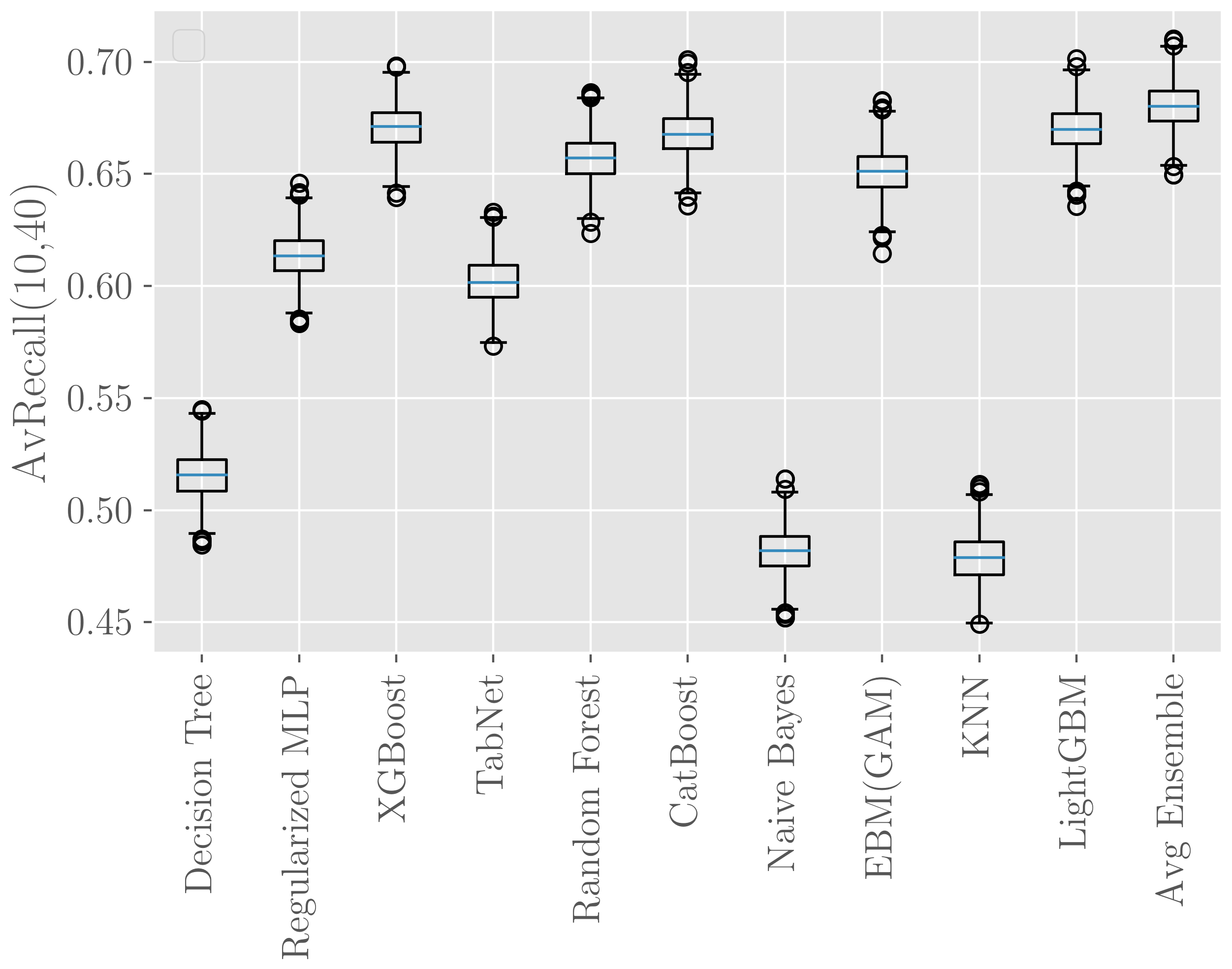}}
\vskip -0.1in
\end{figure}

\begin{figure}[htb!]
\floatconts
  {fig:uncertainty-boxplots-encoders}
  {
  \vspace{-2 em}
  \caption{\small Boxplots with 1000 bootstrap samples for various encoders used with the XGBoost model, generated from the modeling split test set.}}
  {\includegraphics[width=\linewidth]{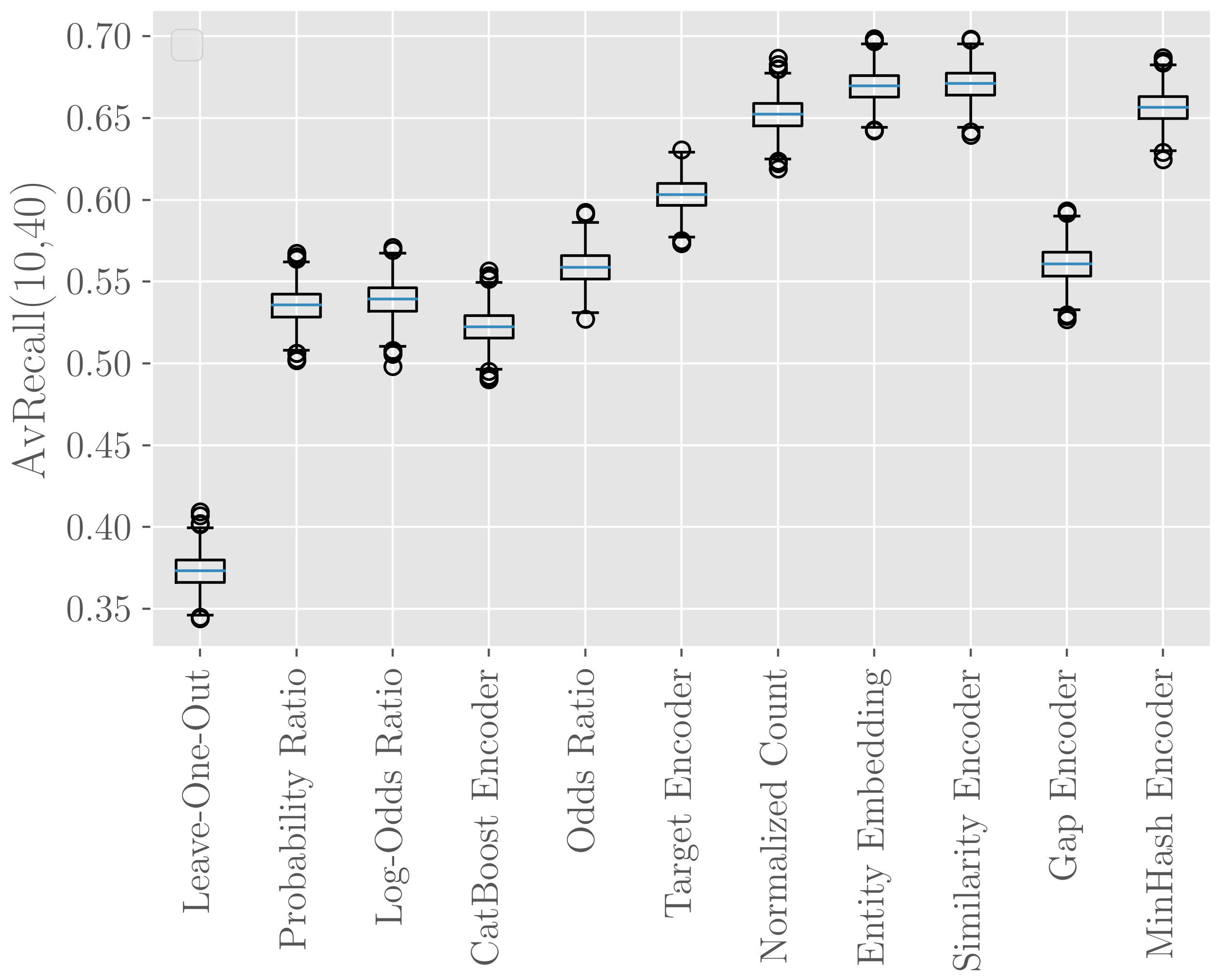}}
\vskip -0.1in
\end{figure}

\section{Improving Performance on Underperforming Cohorts}
We explore two classes of methods to improve performance on underperforming cohorts: data augmentation and algorithmic fairness.
\subsection{Data augmentation}
\label{app:subsec:data-augmentation}
We investigate a simple cohort-level data augmentation scheme that trains a separate model per cohort, trained on training data augmented with $N=10$ copies of all patients from the category of interest. The idea here is to add an extra signal in the training data corresponding to this category.  When we apply this technique to districts, we find that the mean Recall@20 significantly improves for underperforming districts. For instance, the mean recall for all districts with Recall@20 $< 0.3$ changes from $0.175$ to $0.328$, a $87.43\%$ increase, albeit with an increase in effective $k$. This is somewhat similar to increasing the weight of patients from that district in the loss function. Our models are trained on the full modeling split, with results reported on the passive evaluation split. See (\autoref{tab:data_aug}) for detailed results, reported using global thresholds.

Several models (logistic regression, LightGBM, XGBoost, random forest, and regularised MLP)  have the option to use reweighted loss as a  hyperparameter, but the augmentation method was found to do better on underperforming cohorts. We also investigated augmentation with log-inverse frequency (each patient is reweighed by $\log (N/N_d)$ where $N$ is the total number of patients and $N_d$ is the number of patients in that patient's district) but that does not yield any further benefit.  Other data augmentation techniques such as SMOTE \citep{chawla2002smote} did not help either. This may be because performing augmentations with high-dimensional encoded categorical data may not be as useful as with other applications where interpolation is more natural.
\begin{table*}[htb!]
\floatconts
  {tab:data_aug}%
  {\caption{\small Results for data augmentation using $N=10$ copies of the district of interest. The first column shows the max Recall@20 value for the districts aggregated for a particular row.}
  \vspace{-1 em}
  }%
  {%
  \footnotesize
\begin{tabular}{cccccc}
\toprule
\textbf{Threshold} & \textbf{Statistic}& $\mathbf{Recall@20^{\text{(Original)}}}$ & $\mathbf{Recall@20^{\text{(Mitigated)}}}$ & \textbf{\# Districts} &  \textbf{Lift\%} \\
\midrule
0.2 & Mean &          0.080 &           0.400 &           4 &  402.832 \\
    & Std. dev. &          0.093 &           0.437 &           4 &  369.975 \\
\midrule
0.3 & Mean &          0.175 &           0.328 &           9 &   87.316 \\
    & Std. dev. &          0.108 &           0.294 &           9 &  172.990 \\
\midrule
0.4 & Mean &          0.296 &           0.355 &          27 &   20.155 \\
    & Std. dev. &          0.108 &           0.175 &          27 &   61.993 \\
\midrule
0.5 & Mean &          0.404 &           0.437 &          73 &    8.170 \\
    & Std. dev. &          0.109 &           0.145 &          73 &   33.100 \\
\midrule
0.6 & Mean &          0.472 &           0.485 &         133 &    2.638 \\
    & Std. dev. &          0.112 &           0.129 &         133 &   15.088 \\
\midrule
1.0 & Mean &          0.586 &           0.575 &         222 &   -1.834 \\
    & Std. dev. &          0.177 &           0.175 &         222 &   -1.428 \\
\bottomrule
\end{tabular}
}%
\end{table*}

\subsection{Algorithmic fairness}
\label{app:subsec:algorithmic-fairness}

For deployment in sensitive healthcare settings, it is critical that the model is fair across different important cohorts. When we analyze our model performance across cohorts to diagnose this issue, we note the variance in performance across geographies, in particular districts, which is the unit of deployment. While the model does satisfactorily on non-location cohorts, when we look at gender, we note that it does worse on the female cohort.

To improve this, we explore a post-hoc fairness technique inspired by \citep{rodolfa2021empirical}  to equalize performance across these cohorts (\tableref{tab:algo_fair_states} for states and \tableref{tab:algo_fair_underperforming_districts} for districts and \tableref{tab:algo-fair-gender} for gender). The technique involves training a model, changing model scores of patients such that recall is roughly equalized across cohorts on a hold-out set, and then using the same model to evaluate on a test set, using these new cohort-level thresholds. The authors recommend these be forward-split in time. We train on the training set of the modeling split and use global thresholds to compute Recall@20. We then use a combination of the test and val sets to tune cohort-level thresholds and evaluate on the passive evaluation split.

\begin{table*}[htbp]
\centering
\floatconts
  {tab:algo_fair_states}%
  {\caption{\small State-wise analysis of post-hoc threshold adjustment fairness technique, with recall and effective $k$ before and after adjustment. We note an improvement for all states except the best-performing state Uttar Pradesh.}
  \vspace{-1 em}
  }%
{
\footnotesize
\begin{tabular}{ccccccc}
\toprule

\textbf{State} & $\mathbf{Recall@20^{\text{(Original)}}}$ & $\mathbf{Recall@20^{\text{(Mitigated)}}}$ &  \textbf{\#LFUs} &  $\mathbf{k^{\text{(Original)}}}$ & $\mathbf{k^{\text{(Mitigated)}}}$ &  $\mathbf{\Delta}$ \textbf{True +ves} \\
\midrule
Karnataka     &          0.518 &           0.750 &       156.0 &                11.838 &                 27.140 &                   36 \\
Maharashtra   &          0.609 &           0.653 &       244.0 &                14.771 &                 17.850 &                   10 \\
Uttar Pradesh &          0.661 &           0.570 &      1108.0 &                24.986 &                 18.744 &                  -99 \\
West Bengal   &          0.401 &           0.608 &       146.0 &                10.625 &                 24.674 &                   30 \\
\bottomrule
\end{tabular}
}
\end{table*}


\begin{table*}[htbp]
\centering
\floatconts
  {tab:algo_fair_underperforming_districts}
  {\caption{\small Summary performance across districts using the algorithmic fairness technique, with recall and effective $k$ before and after adjustment. We note a very significant increase in performance for underperforming cohorts.}
  \vspace{-1 em}
  }
  {\footnotesize

\begin{tabular}{cccccccc}
\toprule 
\textbf{Threshold} & \textbf{Statistic}& $\mathbf{Recall@20^{\text{(Original)}}}$ & $\mathbf{Recall@20^{\text{(Mitigated)}}}$ &  $\mathbf{k^{\text{(Original)}}}$ & $\mathbf{k^{\text{(Mitigated)}}}$ & \textbf{\# Districts} \\
\midrule
0.2 & Mean &          0.087 &           0.401 &                 6.055 &                 22.595 &          12 \\
    & Std. dev. &          0.077 &           0.297 &                 1.853 &                  9.274 &          12 \\
\midrule
0.3 & Mean &          0.173 &           0.403 &                 8.287 &                 21.247 &          26 \\
    & Std. dev. &          0.098 &           0.253 &                 5.118 &                  9.644 &          26 \\
\midrule
0.4 & Mean &          0.270 &           0.482 &                 9.503 &                 21.511 &          54 \\
    & Std. dev. &          0.118 &           0.230 &                 5.449 &                  8.480 &          54 \\
\midrule
0.5 & Mean &          0.354 &           0.511 &                11.505 &                 21.642 &          96 \\
    & Std. dev. &          0.131 &           0.216 &                 6.143 &                  9.340 &          96 \\
\midrule
0.6 & Mean &          0.403 &           0.530 &                13.031 &                 21.561 &         127 \\
    & Std. dev. &          0.144 &           0.210 &                 6.769 &                  9.373 &         127 \\
\midrule
1.0 & Mean &          0.543 &           0.578 &                18.318 &                 20.671 &         212 \\
    & Std. dev. &          0.213 &           0.208 &                11.095 &                  8.347 &         212 \\
\bottomrule
\end{tabular}

}
\end{table*}

\begin{table*}[htbp]
\centering
\floatconts
  {tab:algo-fair-gender}%
  {\caption{\small Analysis of post-hoc threshold adjustment technique for the gender cohort, with recall and effective $k$ before and after adjustment. We note a modest improvement in performance for the ``Female'' gender.}
  \vspace{-1 em}
  }%
{
\footnotesize
\begin{tabular}{ccccccc}
\toprule

\textbf{State} & $\mathbf{Recall@20^{\text{(Original)}}}$ & $\mathbf{Recall@20^{\text{(Mitigated)}}}$ &  \textbf{\#LFUs} &  $\mathbf{k^{\text{(Original)}}}$ & $\mathbf{k^{\text{(Mitigated)}}}$ &  $\mathbf{\Delta}$ \textbf{True +ves} \\
\midrule
Female      &          0.560 &           0.589 &       582.0 &                16.751 &                 18.754 &                   16 \\
Male        &          0.646 &           0.627 &      1071.0 &                22.332 &                 20.894 &                  -21 \\
Transgender &          0.750 &           0.750 &         1.0 &                22.963 &                 20.741 &                    0 \\

\bottomrule
\end{tabular}
}
\end{table*}

\begin{table*}[htbp]
\centering
\floatconts
   {tab:algo_fair_summary_stats}
  {\caption{\small Summary of inequality measures across states, districts, and gender upon application of the algorithmic fairness technique. We find improvements in the mean and standard deviation of Recall@20 for states and gender, but a worsening of both for districts.}
  \vspace{-1 em}
  }
  {\footnotesize
\begin{tabular}{ccc}
\toprule
\textbf{Metric} & $\mathbf{Recall@20^{\text{(Original)}}}$ & $\mathbf{Recall@20^{\text{(Mitigated)}}}$ \\
\midrule
Overall performance (State level) &           0.616   &            0.606 \\
Gini coefficient (State level) &          0.099 &          0.057  \\
Mean  (State level) &           0.547 &        0.646  \\
Standard deviation  (State level)&        0.114 &	0.078 \\
\midrule
Overall performance (District level) &       0.616 & 0.561 \\
Gini coefficient (District level) &          0.222 &           0.202 \\
Mean  (District level) &        0.543 &      0.578 \\
Standard deviation  (District level)&       0.213 &           0.208 \\
\midrule
Overall performance (Gender) &       0.616 & 0.613 \\
Gini coefficient  (Gender)      &          0.065 &           0.055               \\
Mean    (Gender)    &          0.652 &           0.655               \\
Standard deviation (Gender)       &          0.095      &  0.084             \\
\bottomrule

\end{tabular}
}
\end{table*}

We get an increase in model performance across underperforming cohorts for all covariates analyzed -- states, districts, and gender. We see all states but the best-performing show an improvement and see particularly significant improvements for very poorly performing districts, with a $132\%$ increase in mean recall for districts with Recall@20 $< 0.3$. We only use districts that have at least one LFU patient in train, union of val and test, and passive evaluation sets, which reduces the total number of districts to 212 from 224.  Our analysis for gender shows that the application of the fairness technique reduces variation in performance between females and males. Overall results for all methods are summarized in \tableref{tab:algo_fair_summary_stats}.

\end{document}